# Data Valuation for Vertical Federated Learning: A Model-free and Privacy-preserving Method

**Xiao Han**
School of Information Management and Engineering,
Shanghai University of Finance and Economics, Shanghai, 200433, CHINA
xiaohan@mail.shufe.edu.cn

**Leye Wang**
Key Lab of High Confidence Software Technologies (Peking University), Ministry of Education
School of Computer Science, Peking University, Beijing, 100871, CHINA
leyewang@pku.edu.cn

**Junjie Wu**
Key Laboratory of Data Intelligence and Management (Beihang University),
Ministry of Industry and Information Technology
School of Economics and Management, Beihang University, Beijing 100191, CHINA
wujj@buaa.edu.cn

**Xiao Fang**
Lerner College of Business and Economics,
University of Delaware, Newark, DE 19716 U.S.A.
xfang@udel.edu

**Abstract**

Vertical Federated learning (VFL) is a promising paradigm for predictive analytics, empowering an organization (*i.e.*, *task party*) to enhance its predictive models through collaborations with multiple data suppliers (*i.e.*, *data parties*) in a decentralized and privacy-preserving way. Despite the fast-growing interest in VFL, the lack of effective and secure tools for assessing the value of data owned by data parties hinders the application of VFL in business contexts. In response, we propose *FedValue*, a privacy-preserving, task-specific but model-free data valuation method for VFL, which consists of a data valuation metric and a federated computation method. Specifically, we first introduce a novel data valuation metric, namely *MShapley-CMI*. The metric evaluates a data party's contribution to a predictive analytics task without the need of executing a machine learning model, making it well-suited for real-world applications of VFL. Next, we develop an innovative federated computation method that calculates the MShapley-CMI value for each data party in a privacy-preserving manner. Extensive experiments conducted on six public datasets validate the efficacy of FedValue for data valuation in the context of VFL. In addition, we illustrate the practical utility of FedValue with a case study involving federated movie recommendations.

# Data Valuation for Vertical Federated Learning: A Model-free and Privacy-preserving Method

## 1 INTRODUCTION

Owing to the sheer volume of data accumulated in organizations, predictive analytics has become increasingly vital for businesses across industries (Agarwal and Dhar, 2014; Rai, 2017). It analyzes data with machine learning techniques and makes predictions about future events or unknown facts, which in turn supports organizational decision makings effectively. As a result, the global market size of predictive analytics had more than doubled from $4.2 billion in 2017 to $9.1 billion in 2021[1], and is expected to continue growing to $28.1 billion by 2026[2]. Recognizing the value of predictive analytics, Information Systems (IS) researchers have developed predictive analytics methods to solve a variety of critical business and societal problems, including social network analytics (Fang and Hu, 2018), mobile app recommendations (He et al., 2019), and healthcare analytics (Yu et al., 2023). To maximize the value of predictive analytics, it is important to employ both internal data within an organization and external data outside the organization (Kitchens et al., 2018; Lukyanenko et al., 2019). For example, Lu et al. (2023) conduct a field experiment on a microloan platform and find that combining internal data (*e.g.*, loan amount) within the platform and external data (*e.g.*, loan applicants' cellphone usage data) outside the platform can significantly enhance the accuracy of default predictions, in comparison to using internal data solely. Wang et al. (2021a) develop a predictive analytics method that utilizes multi-source data collected from within and outside a firm and demonstrate that the method can model financial risks more comprehensively and improve the accuracy of financial risk predictions remarkably. Although these studies show the great potential for using external data for predictive analytics, they overlook two obstacles that hinder the practical utilization of external data (Kitchens et al., 2018; Liu et al., 2021a). First, owing to significant concerns about

[1] https://www.statista.com/statistics/819415/worldwide-predictive-analytics-market-size/

[2] https://www.marketsandmarkets.com/Market-Reports/predictive-analytics-market-1181.html

data privacy, several data regularizations and laws, such as General Data Protection Regulation[3] in Europe and Data Security Act[4] in China, have been enacted to prohibit sharing raw data among organizations without proper authorization (Xu and Zhang, 2022). Second, because data are one of the most valuable assets for organizations, data holders are unwilling to share their data with other organizations without appropriate incentives (e.g., monetary compensation) (Wang et al., 2021b).

Federated learning (FL) is an emerging paradigm towards privacy-preserving collaborative machine learning for predictive analytics (Bi et al., 2023; Yang et al., 2019; Konečnỳ et al., 2016). Generally, FL enables its participating organizations to collaboratively train a predictive analytics model without sharing raw data among them. More specifically, each participating organization learns the parameters of a predictive analytics model from its own data locally and only transfers learned parameters to other participating organizations for collaborative model learning, thereby preserving the privacy of raw data. Depending on how data are distributed among participating organizations, there are two types of FL: horizontal FL (HFL) and vertical FL (VFL) (Yang et al., 2019). For HFL, local data at different participating organizations have the same features but describe different entities (*e.g.*, customers) (Wang et al., 2020; Wei et al., 2020). Comparatively, for VFL, local data at different participating organizations describe the same set of entities with different features (Wu et al., 2020b; Wang et al., 2019).

Our study focuses on VFL, which is particularly suitable for business applications. Figure 1 illustrates an application of VFL to credit risk prediction. As shown, a financial institute collaborates with a mobile service provider and an online shopping platform to build a credit risk prediction model by analyzing its internal data about credit applicants as well as these applicants' mobile usage and online shopping data (*i.e.*, external data). In general, if a party (namely task party, such as a financial institute) needs to train a predictive analytics model by exploiting additional features about its data samples, it can invite other parties (namely data parties, such as a mobile service provider) who possess these additional features to launch a VFL task.

[3] `https://gdpr.eu/eu-gdpr-personal-data`
[4] `http://www.npc.gov.cn/npc/c30834/202106/7c9af12f51334a73b56d7938f99a788a.shtml`

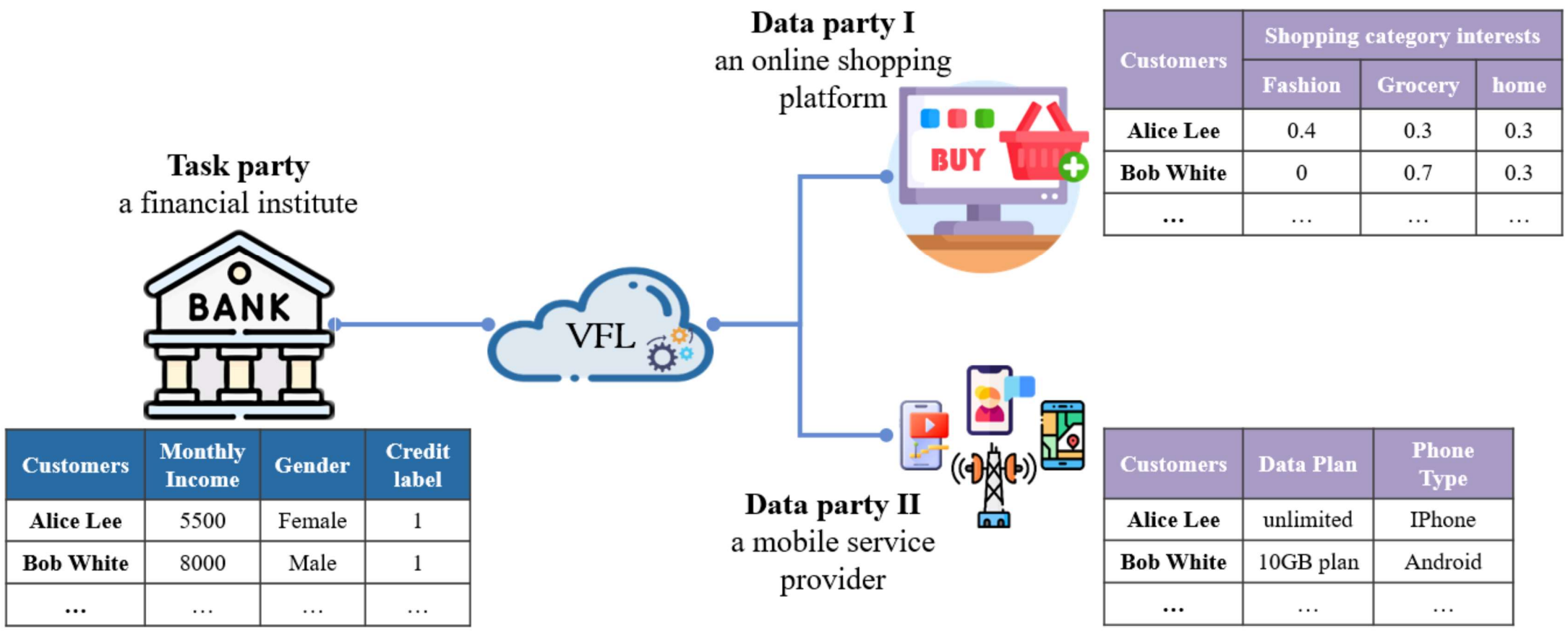

**Figure 1. An application of VFL. A financial institute as a task party collaborates with two data parties (an online shopping platform and a mobile service provider) to train a credit risk prediction model using internal data within the financial institute and external data from the data parties.**

Although VFL offers a viable solution to overcome the privacy obstacle of integrating internal and external data for predictive analytics, how to select suitable collaborators from many external data suppliers and determine appropriate incentives to encourage high-quality external data suppliers remains under-explored (Kairouz et al., 2021). In response, this study tackles the selection and incentive obstacles by developing a novel data valuation method for VFL. To achieve a successful VFL task, a task party must carefully select the appropriate data parties and offer them proper incentives. Data valuation quantifies the contribution of a data party's external data to a VFL task. Therefore, an effective data valuation method helps the task party select data parties that contribute most to the VFL task. Moreover, it aids the task party in determining suitable incentives for the selected data parties and establishing a contribution-based incentive mechanism. This mechanism is crucial for creating a fair and sustainable data marketplace for predictive analytics (Yang et al., 2019).

Designing an effective data valuation method for VFL is non-trivial due to the following challenges. First, an effective data valuation method needs to be task-specific and model-free. Data should be valuated within the context of a particular task because the same data could make different contributions to different tasks (Wang and Strong, 1996; Wang et al., 2021b). For example, the contributions of customers' mobile usage data to the task of credit risk prediction might be different from their contributions to the task of customer retention. Furthermore, an effective data valuation method should be independent of machine learning models. A data valuation method is employed by a task party to evaluate and select data parties for a VFL task. At this early stage, task and data parties are aware of the specific task but have not decided on the particular machine learning model for the task. For example, task and data parties understand that a credit risk prediction task is to be performed but have not determined which machine learning model (*e.g.*, logistic regression or XG-Boost) to use. Existing data valuation methods fail to address this challenge for they are either task-independent or model-dependent. Several task-independent metrics have been proposed to valuate data according to their general properties, including completeness, precision, uniqueness, and timeliness (Wang and Strong, 1996; Batini et al., 2009); these metrics, however, are not tailored to assess the contribution of data to a specific task. Although task-specific data valuation metrics, such as the SHAP value (Lundberg and Lee, 2017), consider the properties of a specific task, they are model-dependent. To calculate the SHAP value of external data, it needs to train a specific machine learning model using that data. In practice, a data party may be hesitant to share their data for model training until they are certain of being selected and adequately compensated with appropriate incentives.

Second, there might be overlapping or correlated features between the internal data of the task party and the external data from a data party. This phenomenon is known as the "substitution effect", where these same or correlated features contribute in a comparable manner to a predictive task. Therefore, an effective data valuation method should take the substitution effect into account and determine appropriate incentives for data parties based on their actual contributions to a VFL task initiated by the task party. Third, a data valuation method must comply with the privacy-preserving requirement. That is, a data valuation method should adhere to data protection regulations and laws, ensuring that no raw data can be leaked from any party involved. However, to assess the contribution of external data, it is essential to have relevant information about the data. For example, we need to know the features of external data to quantify the substitution effect discussed above. As a result, how to acquire useful information about external data without violating the privacy-preserving re-quirement, is another challenge for developing an effective data valuation method.

To address these challenges, we propose a privacy-preserving, task-specific but model-free data valuation method for VFL tasks. In particular, premised in cooperative game theory and information theory, we model a VFL task as a master-managed cooperative prediction game and propose a task-specific but model-free data valuation metric that accounts for the substitution effect. We then develop a federated computation method that assesses the contribution of external data from a data party to a VFL task using the proposed metric. Our data valuation method is privacy-preserving in that data are securely stored at their owning data parties. Moreover, it can effectively verify the trustworthiness of results produced by a server in a VFL task, thereby safeguarding against potential malicious attacks by the server intended to manipulate results. We conduct extensive experiments on six real-life datasets to demonstrate the effectiveness of our method in valuating data for VFL tasks, along with its computational efficiency. We also show the practical value of our method in the context of a federated recommendation task.

## 2 RELATED WORK

Two streams of research are closely related to our study: federated learning (FL) and data valuation. We review each of them and highlight the key novelties of our study.

### Federated Learning

FL enables decentralized collaborative machine learning by utilizing data distributed across multiple parties while ensuring that no party's data are leaked to any others (Yang et al., 2019). Due to the growing concerns about data privacy, FL has garnered substantial attention from both academia and industry (Yang et al., 2019). An important domain that embraces FL for privacy-preserving predictive analytics is business. In this vein, Tan et al. (2020) deploy a federated recommender system for content recommendations. Liu et al. (2021a) develop a FL framework that enables multiple banks to collaboratively build anti-money laundering models without sharing data among these banks. FL has also been adopted for predictive analytics in the domain of healthcare. For example, Kaissis et al.

(2021) develop a framework of federated deep neural networks by exploiting medical image data stored at multiple institutions for privacy-preserving healthcare analytics such as pediatric chest radiography classification. Warnat-Herresthal et al. (2021) leverage blood transcriptomes and chest X-ray image data distributed across different medical centers and hospitals to build federated sequential deep neural networks for disease (*e.g.*, COVID-19, tuberculosis, and leukemias) prediction.

The majority of FL studies design privacy-preserving and decentralized machine learning algorithms that allow data to be kept locally and only exchange model parameters or intermediate results for collaborative training. FedSGD and FedAvg are two pioneering FL algorithms, which are developed for deep neural networks (McMahan et al., 2017). Specifically, under the framework of FedSGD, each participation party trains its local neural network model using its own data and sends model gradients to a trusted server. The server then aggregates these gradients to create a comprehensive model without accessing the raw data from individual parties, ensuring privacy and confidentiality. Built on FedSGD, FedAvg further improve its computational efficiency and privacy protection. Based on these pioneering algorithms, recent studies have extended FL to other widely-used predictive analytics methods, including logistic regression (Hardy et al., 2017; Hu et al., 2019), decision trees (Cheng et al., 2019; Wu et al., 2020b), and matrix factorization (Chai et al., 2020). In addition, some other studies focus on enhancing FL algorithms by tackling challenges such as non-i.i.d. (non-independent and identically distributed) data among different participating parties or optimizing the communication efficiency among these parties (Sattler et al., 2019; Wu et al., 2020a). Different from these studies, our paper tackles the data valuation problem, which quantifies the contribution of external data to a predictive analytics task. Data valuation is essential to a successful FL application because it helps the task party select data parties that contribute most to the application and facilitates the determination of appropriate incentives offered to them.

## Data Valuation

IS researchers have pioneered the data valuation research and proposed a number of data quality metrics from theoretical and operational perspectives (Wang and Strong, 1996; Batini et al., 2009; Wang et al., 2021b). These metrics evaluate the quality of data based on their intrinsic characteristics, including *accuracy*, *completeness*, *consistency*, and *timeliness* (Batini et al., 2009). In addition, some studies consider the contextual nature of data quality and define system-dependent data quality measurements such as *relevancy*, *accessibility*, *understandability*, *traceability*, and *usage* (Wang and Strong, 1996; Wang et al., 2021b). Data quality metrics provide general and high-level evaluations of data value, but they are not tailored for data valuation in the context of predictive analytics.

In response, several data importance metrics have been proposed to assess the contribution of data in a predictive analytics task. Some of these metrics are developed for particular machine learning models. For example, *variable importance* (VI) and *Gini importance* (GI) are data importance measurements developed for predictive analytics with tree-based models (Breiman, 2001). In addition, for a predictive analytics task employing a linear model (*e.g.*, logistic regression), the co-efficient of a feature indicates its contribution to the task (Fisher et al., 2019a). Some other data importance metrics, such as *permutation importation* (PI) (Fisher et al., 2019a), *LIME* (Ribeiro et al., 2016), and *SHAP* (Lundberg and Lee, 2017; Jia et al., 2019), can work with different machine learning models, but evaluation results of these metrics depend on the particular machine learning model used in an evaluation. Specifically, PI assesses the importance of a feature by examining how the performance of a machine learning model varies when different values of the feature are used (Altmann et al., 2010). LIME approximates a complex machine learning model with an interpretable surrogate model (*e.g.*, a linear model) and then uses the surrogate model to evaluate feature importance (Ribeiro et al., 2016). By treating each feature as a game player and the prediction outcome of a predictive analytics task as the payout of the game, SHAP allocates the payout to each feature based on its Shapley value and measures the importance of a feature using its allocated payout. Also built on the Shapley Value, Jia et al. (2019) design an efficient method to measure the contribution of data samples to a predic-tive analytics task by proposing several approximation techniques to compute Shapley Values. Data importance metrics are developed for centralized predictive analytics tasks, where data are collected and analyzed in a centralized manner. Therefore, they are not suitable for distributed predictive analytics tasks, where data are distributed across various task and data parties and the privacy of local data at each party must be protected.

To address the data valuation challenges encountered in distributed predictive analytics, several pioneering studies have developed data valuation methods within the framework of FL (Yu et al., 2020). In particular, Wang et al. (2019) propose the first privacy-preserving data valuation method for distributed predictive analytics. The method utilizes the Shapley Value to assess the contribution of data samples under the framework of HFL or gauge the importance of data features in the context of VFL. Following (Wang et al., 2019), subsequent studies mainly target HFL and focus on reducing the cost of computing the Shapley value (Wang et al., 2020; Wei et al., 2020; Liu et al., 2021b). Although VFL has a large number of business applications, data valuation for VFL has not received sufficient research attention (Liu et al., 2022). VerFedSV is such an attempt that reduces the communication costs among participating parties in a VFL task, thereby improving the overall efficiency of data valuation for VFL (Fan et al., 2022).

As discussed before, an effective data valuation method needs to be task-specific, model-free, and privacy-preserving, and it should take the substitution effect into consideration. Our literature review suggests that existing data valuation methods lack in some of these aspects, as summarized in Table 1. Although data quality metrics are model-free, they are too general and not tailored for a specific predictive analytics task. Data importance metrics, on the other hand, are developed for a specific predictive analytics task but they are model-dependent. For example, *Gini importance* are designed for tree-based models (Breiman, 2001) whereas *SHAP* (Lundberg and Lee, 2017) assesses the importance of a feature by executing a specific machine learning model. Furthermore, both data quality and data importance metrics fail to consider the substitution effect, and they are not designed to protect the privacy of data distributed across various parties. Existing data valuation methods for HFL or VFL are developed for a specific predictive analytics task under the framework of FL. Therefore, they are task-specific and privacy-preserving. However, these methods evaluate the contribution of data through the execution of a particular machine learning model (Wei et al., 2020). Thus, they are model-dependent. Moreover, these methods divide the contribution of an overlapping attribute evenly among its owners (Fan et al., 2022). Consider an example of a task party and two data parties, each of which owns an overlapping attribute A. Suppose the contribution of the attribute to a predictive analytics task is $1$ and these methods then attribute the value of attribute A owned by each party as $1/3$. Such a way of valuation overlooks the substitution effect between internal data owned by the task party and external data held by data parties. Recall that the problem context of this study is that a task part seeks external data from data parties to use in a predictive analytics task. Given that the task party already owns attribute A, the value of attribute A held by each data party should be $0$ to the task party, rather than $1/3$. In recognition of the deficiencies of the existing methods, we develop a task-specific, model-free, and privacy-preserving data valuation method that takes the substitution effect into account.

**Table 1. Comparison between our method and existing data valuation methods.**

| Method | Task-specific | Model-free | Considering substitution effect | Privacy-preserving |
|---|---|---|---|---|
| Data quality metrics, *e.g.*, Wang and Strong (1996) | | √ | | |
| Data importance metrics, *e.g.*, Ribeiro et al. (2016) | √ | | | |
| Data valuation methods for HFL, *e.g.*, Wei et al. (2020) | √ | | | √ |
| Data valuation methods for VFL, *e.g.*, Wang et al. (2019) | √ | | | √ |
| Our method | √ | √ | √ | √ |

## 3 PROBLEM FORMULATION

We consider a scenario where a task party collaborates with data parties to train a predictive analytics model. In this scenario, the task party seeks to build the model and data parties provide external data with complementary features for model training.

**Definition 1** (Task Party). *A task party $t$ holds a set of data samples, each of which is identified by its ID $i \in \mathcal{I}_t$, where $\mathcal{I}_t$ represents the IDs of all data samples held by the task party. Each data sample $t[i]$ is characterized by its feature vector $\boldsymbol{x}_{t[i]} \in \mathbb{R}^{|F_t|}$ and task label $y_{t[i]} \in \mathbb{R}$, where $F_t$ denotes the features held by the task party. Accordingly, data held by the task party can be represented as $(\boldsymbol{X}_t, \boldsymbol{Y}_t)$, where $\boldsymbol{X}_t \in \mathbb{R}^{|\mathcal{I}_t| \times |F_t|}$ is the feature matrix, each row of which is $\boldsymbol{x}_{t[i]}$, and $\boldsymbol{Y}_t \in \mathbb{R}^{|\mathcal{I}_t| \times 1}$ denotes the label vector, each element of which is $y_{t[i]}$.*

**Definition 2** (Data Party). *A data party $d$ possesses a set of data samples, each of which is identified by its ID $i \in \mathcal{I}_d$, where $\mathcal{I}_d$ denotes the IDs of all data samples at the data party. Each data sample $d[i]$ is described by its feature vector $\boldsymbol{x}_{d[i]} \in \mathbb{R}^{|F_d|}$, where $F_d$ represents the features held by the data party. We use feature matrix $\boldsymbol{X}_d \in \mathbb{R}^{|\mathcal{I}_t| \times |F_d|}$ to represent data owned by the data party, and each row of the matrix is $x_{d[i]}$.*

Figure 2 illustrates these two definitions. As shown, a task party initiates a Vertical FL (VFL) task by inviting selected data parties with suitable incentives. The objective of this task is to train a predictive analytics model by utilizing both internal data held by the task party and external data owned by the selected data parties. A VFL task can be identified by its associated task label $Y_t$. For example, the task label of credit risk prediction is whether a customer will default whereas the task label of customer retention is whether a customer will churn. To successfully launch a VFL task, two problems need to be solved. First, common data samples $\mathcal{I}$ among task and data parties must be identified in a privacy-preserving manner, where $\mathcal{I} = (\bigcap_{d \in \mathcal{D}} \mathcal{I}_d) \cap \mathcal{I}_t$ and $\boldsymbol{\mathcal{D}}$ denotes the set of all available data parties. This problem has been successfully solved by the privacy-preserving data alignment method (Cheng et al., 2019). Second, a task party needs to select appropriate data parties and provide suitable compensations to the selected data parties. To this end, a task party should have the capability to assess the contribution of external data from a data party to the VFL task. Accordingly, we define the data valuation problem for VFL below and solve the problem in this study.

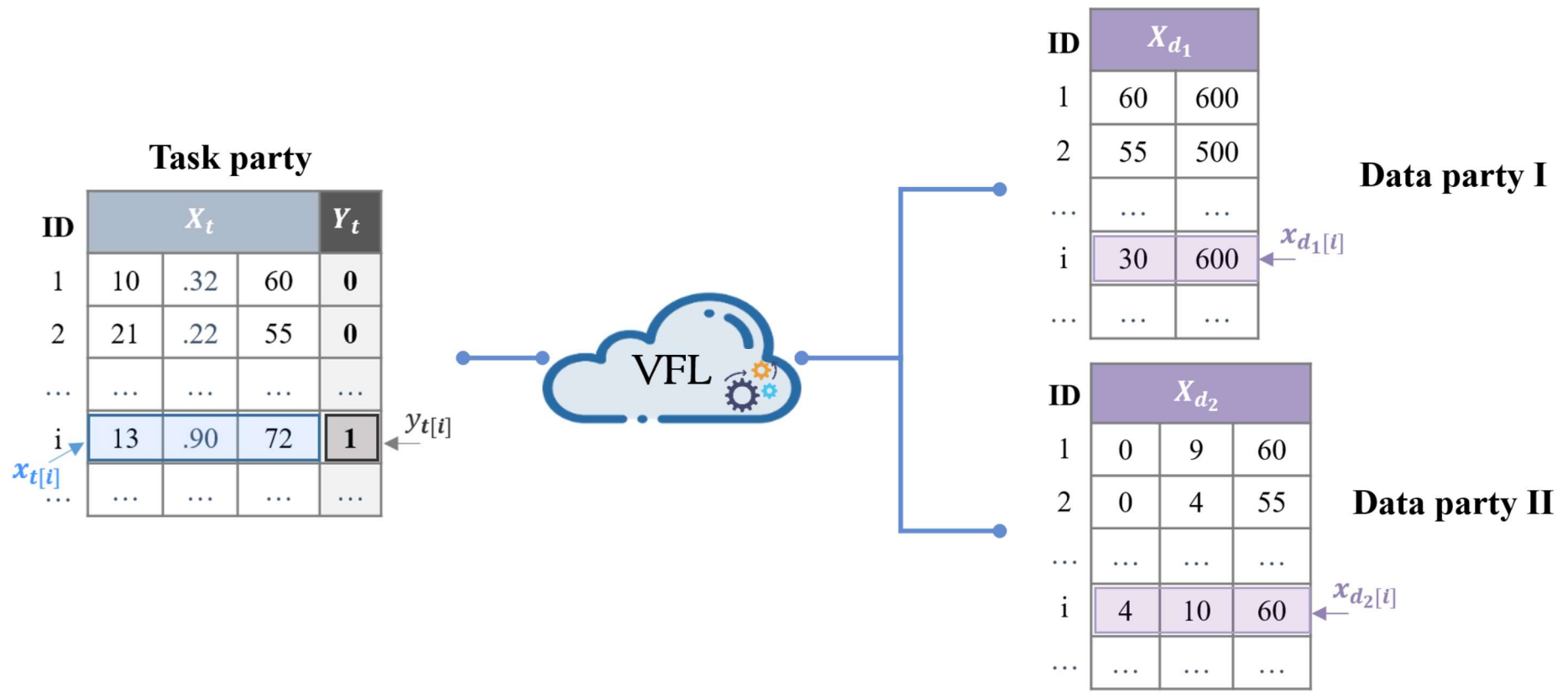

**Figure 2. An illustrative example of data held by task and data parties in a VFL task.**

**Definition 3** (Data Valuation Problem for VFL)**.** *Given a VFL task, a task party $t$ and its internal data, a set of data parties $d \in \mathcal{D}$, each of which holds its own external data, the objective of the problem is to design a task-specific, model-free, and privacy-preserving method that evaluates the contribution of external data owned by a data party to the VFL task, while taking into account the substitution effect among data held by task and data parties.*

## 4 METHOD

In this section, we propose *FedValue*, a privacy-preserving, task-specific but model-free method, for data valuation in a VFL task. Concretely, FedValue consists of two components: a data valuation metric and a federated computation method. Premised in cooperative game theory and information theory, we design a task-specific but model-free data valuation metric that accounts for the substitution effect among data owned by task and data parties. Armed with the data valuation metric, we further develop a federated computation method that assesses the contribution of data owned by a data party to a VFL task in a privacy-preserving manner.

### Data Valuation Metric

In a VFL task, the task party seeks to improve its business model by offering incentives to data parties for a cooperative prediction. Thus, it is natural to model a VFL task as a cooperative game. Along this line, we begin by revisiting *cooperative game* and its core solution concept – *Shapley*

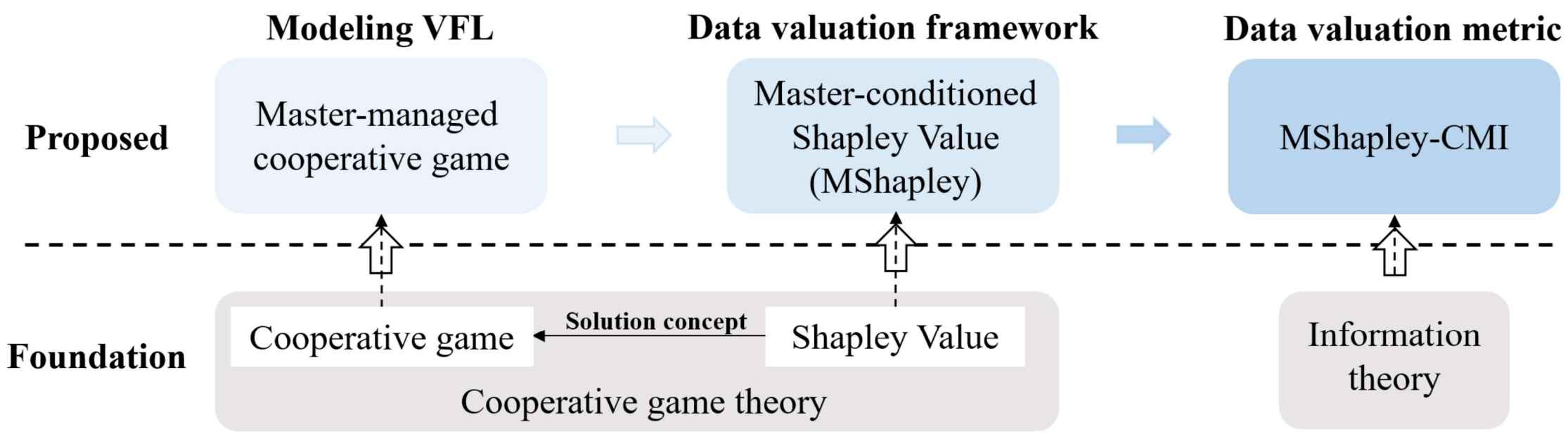

**Figure 3. The proposed data valuation metric and its theoretic foundations.**

*Value*. We then model a VFL task as a *master-managed cooperative prediction game* and propose the *Master-conditioned Shapley Value* that considers the substitution effect among data owned by task and data parties. On this basis, we further design an information-theoretic data valuation metric named *MShapley-CMI* that is task-specific but model-free. Fig. 3 illustrates the process of designing the metric.

**Preliminaries: Cooperative Game and Shapley Value**

Generally speaking, players participating in a cooperative game form coalitions by making binding agreements, so that they can work together to achieve a common goal and receive benefits from the cooperation. A cooperative game is typically modeled as a pair $(\mathbb{P}, v)$, where $\mathbb{P}$ denotes the set of players in the game, $v$ is the characteristic function, and $v(S)$ defines the real-numbered value generated by coalition $S \subseteq \mathbb{P}$. Shapley value, a core solution concept in cooperative game theory, determines how to allocate the value generated by a coalition among its participants (Shapley, 1953). Specifically, the Shapley value of player $j \in \mathbb{P}$ is defined as:

$$\varphi_j = \sum_{S \subseteq (\mathbb{P} \setminus \{j\})} \frac{|S|!(|\mathbb{P}| - |S| - 1)!}{|\mathbb{P}|!} \left(v(S \cup \{j\}) - v(S)\right). \tag{1}$$

In this equation, $v(S \cup \{j\}) - v(S)$ computes $j$'s *marginal contribution* to the game if the player joins the game right after coalition $S$. The first term of the equation, $(|S|)!(|\mathbb{P}| - |S| - 1)!/|\mathbb{P}|!$, gives the probability that player $j$ joins the game right after coalition $S$. Thus, Eq. (1) calculates the

Shapley value of player $j$ as the player's *expected marginal contribution* to the game.

Being regarded as a fair and justifiable value allocation method, Shapley value has been widely adopted in various applications including the task of advertising attribution (Singal et al., 2022), the game of greenhouse gas emissions (Gopalakrishnan et al., 2021), and the estimation of feature importance in centralized machine learning (Lundberg and Lee, 2017). Shapley value has several desirable properties (Shapley, 1953):

- **Efficiency**: $\sum_{j \in \mathbb{P}} \varphi_j = v(\mathbb{P})$.
- **Symmetry**: Consider two players $j, j' \in \mathbb{P}$ such that for any $S \subseteq \mathbb{P} \setminus \{j, j'\}$, $v(S \cup \{j\}) - v(S) = v(S \cup \{j'\}) - v(S)$, then $\varphi_j = \varphi_{j'}$.
- **Null-player**: If a player $j \in \mathbb{P}$ does not add value to any coalition, *i.e.*, for any $S \subseteq \mathbb{P} \setminus \{j\}$, $v(S \cup \{j\}) = v(S)$, then $\varphi_j = 0$.

**Master-managed Cooperative Prediction Game for VFL**

Prior data valuation methods model a Horizontal FL (HFL) task or a Vertical FL (VFL) task as a cooperative prediction game denoted by $(\mathbb{P}, v)$, where $\mathbb{P}$ comprises all the parties participating in the task (Wang et al., 2019; Wei et al., 2020; Fan et al., 2022). These methods then adopt the Shapley value defined in Eq. (1) to assess the contribution of each party to the task. One limitation of prior data valuation methods is that they treat the task party and each data party indifferently. In reality, the task party initiates a VFL task as a master and recruits data parties with complementary data to improve prediction performance. More severely, prior data valuation methods ignore the substitution effect between internal data owned by the task party and external data held by a data party. Consequently, these methods fail to assess the proper value of a data party to a VFL task. We illustrate this limitation using the following example.

**Example 1.** *Consider a VFL task involving a task party $t$ and a data party $d$, where the feature matrix $\boldsymbol{X}_d$ of $d$ is identical to the feature matrix $\boldsymbol{X}_t$ of $t$. By applying the Shapley value defined in Eq.* (1)*, prior data valuation methods assign the value of $d$ as $\varphi_d = \frac{v(\{t,d\})}{2} = \frac{v(\boldsymbol{X}_d)}{2}$. The derivation* *of $\varphi_d$ is provided in Appendix A. In reality, given $\boldsymbol{X}_t$ already held by task party $t$ and $\boldsymbol{X}_t = \boldsymbol{X}_d$, data party $d$ provides no complementary data to the VFL task. As a result, the value of data party $d$ should be 0, i.e., $\varphi_d = 0$. Prior data valuation methods fail to valuate the contribution of data party $d$ properly because they overlook the substitution effect between $\boldsymbol{X}_t$ and $\boldsymbol{X}_d$.*

To address the limitations of prior data valuation methods, we model a VFL task as a *master-managed cooperative prediction game*, and propose a valuation formula that considers the substitution effect.

**Definition 4 (Master-managed Cooperative Prediction Game for VFL).** *A master-managed cooperative prediction game for VFL is defined as a triplet $(t, \boldsymbol{\mathcal{D}}, v)$, where $t$ represents a master player (i.e., task party), $\boldsymbol{\mathcal{D}}$ consists of ordinary players (i.e., data parties), $v$ is the characteristic function of the game, and task and data parties are respectively defined in Definitions 1 and 2. All the players $\mathbb{P} = \{t\} \cup \boldsymbol{\mathcal{D}}$ work together to accomplish effective predictions on task label $Y_t$. Characteristic function $v(S)$ measures the effectiveness of coalition $S$ in predicting $Y_t$, where $S$ is a coalition of task and data parties, i.e., $t \in S \subseteq \mathbb{P}$.*

The differentiation of the task party from data parties in the game defined above lays the foundation for our data valuation metric that takes the substitution effect into account. Concretely, the *MShapley value (Master-conditioned Shapley Value)* of a party in a master-managed cooperative prediction game for VFL is given by,

$$\begin{cases} \varphi_t = v(\{t\}), & \text{(2a)} \\ \varphi_d = \sum\limits_{D \subseteq (\boldsymbol{\mathcal{D}} \setminus \{d\})} \dfrac{(|D|)!(|\boldsymbol{\mathcal{D}}| - |D| - 1)!}{|\boldsymbol{\mathcal{D}}|!} \left(v(S \cup \{d\}) - v(S)\right), \forall\, d \in \boldsymbol{\mathcal{D}}. & \text{(2b)} \end{cases}$$

where $\varphi_t$ denotes the MShapley value of task party $t$, $\varphi_d$ is the MShapley value of data party $d$, and $S = \{t\} \cup D$. Equation (2) prioritizes task party $t$ by fixing it as the first participant in any coalition. As a result, the contribution of $t$ to the prediction game does not depend on any other party and $\varphi_t = v(\{t\})$. Because $t$ is the first participant in any coalition, the contribution of data party $d$ to the prediction game depends on not only $t$ but also other data parties joining the game before $d$. Consequently, given that $d$ joins the game right after coalition $S = \{t\} \cup D$, $v(S \cup \{d\}) - v(S)$ calculates $d$'s *marginal contribution* to the game. In other words, data party $d$ makes contributions to the prediction game if and only if it provides additional useful data to the prediction, given internal data owned by task party $t$ and external data held by other data parties joining the game before $d$. Therefore, Equation (2) considers the substitution effect between internal data owned by the task party and external data held by a data party by fixing $t$ as the first participant of coalition. The term $(|D|)!(|\boldsymbol{\mathcal{D}}| - |D| - 1)!/|\boldsymbol{\mathcal{D}}|!$ in Equation (2) gives the probability that data party $d$ joins the game right after coalition $S = \{t\} \cup D$. Hence, Eq. (2) computes $\varphi_d$ as the expected marginal contribution of data party $d$ by averaging over all possible coalition perturbations with $t$ always fixed as the first participant.

**MShapley-CMI Metric**

Given the MShapley value defined in Equation (2), the remaining task for designing our data valuation metric is to formulate an appropriate characteristic function $v$ that needs to be model-free and task-specific. Recall that coalition $S = \{t\} \cup D$ consists of the task party $t$, which owns feature matrix $\boldsymbol{X}_t$ and labels of its data samples $Y_t$, and a set of data parties $D$, each of which holds its own feature matrix $\boldsymbol{X}_{d_k}$ where $d_k \in D$ and $k = 1, 2, \ldots, |D|$. Let $\boldsymbol{X}_D = (\boldsymbol{X}_{d_1}, \cdots, \boldsymbol{X}_{d_k} \cdots, \boldsymbol{X}_{d_{|D|}})$ denote the aggregated feature matrix owned by $D$ and $\boldsymbol{X}_S = (\boldsymbol{X}_t, \boldsymbol{X}_D)$ be the aggregated feature matrix held by $S$. To design a model-free and task-specific data valuation metric, we measure the value of $\boldsymbol{X}_S$ in predicting task label $Y_t$ based on information theory. Our basic idea is to evaluate this value as the reduced uncertainty in predicting $Y_t$ due to $\boldsymbol{X}_S$.

We first consider the prediction of task label $Y_t$ based on the prior knowledge about $Y_t$ solely, *i.e.*, the distribution $p(Y_t)$ derived from labels of data samples $Y_t$ held by the task party. Premised in information theory, the uncertainty of this prediction can be measured by the *entropy* of $Y_t$ as

$$H(Y_t) = -\sum_{y_t \in \mathcal{Y}_t} p(y_t) \log p(y_t), \tag{3}$$

where $\mathcal{Y}_t$ represents the set of all possible values of task label $Y_t$. In this equation, $p(y_t) = p(Y_t = y_t)$ denotes the probability of $y_t$ being observed, which can be learned from $Y_t$. The uncertainty of predicting $Y_t$ after introducing $\boldsymbol{X}_S$ can then be computed as the conditional entropy of $Y_t$ on $\boldsymbol{X}_S$, which is given by

$$H(Y_t|\boldsymbol{X}_S) = -\sum_{\boldsymbol{x}_S \in \mathcal{X}_S} p(\boldsymbol{x}_S) \sum_{y_t \in \mathcal{Y}_t} p(y_t|\boldsymbol{x}_S) \log p(y_t|\boldsymbol{x}_S), \tag{4}$$

where $\mathcal{X}_S$ denotes the set of all possible values for the features held by coalition $S$, $p(\boldsymbol{x}_S)$ is the probability of feature value $\boldsymbol{x}_S$ being observed in $\boldsymbol{X}_S$, and $p(y_t|\boldsymbol{x}_S)$ represents the conditional probability of $y_t$ given $\boldsymbol{x}_S$.

Given coalition $S$ and its feature matrix $\boldsymbol{X}_S$, the characteristic function $v(S)$ in Eq. (2b) computes the value of $\boldsymbol{X}_S$ as the reduced uncertainty in predicting $Y_t$ attributed to $\boldsymbol{X}_S$. Accordingly, we have

$$v(S) = H(Y_t) - H(Y_t|\boldsymbol{X}_S) = I(\boldsymbol{X}_S; Y_t) = I(\boldsymbol{X}_t\boldsymbol{X}_D; Y_t), \tag{5}$$

where $I(\boldsymbol{X}_S; Y_t)$ is the mutual information between $\boldsymbol{X}_S$ and $Y_t$ (Brown et al., 2012). Consequently, the marginal contribution of data party $d$ in Eq. (2b) is:

$$v(S \cup \{d\}) - v(S) = I(\boldsymbol{X}_d\boldsymbol{X}_t\boldsymbol{X}_D; Y_t) - I(\boldsymbol{X}_t\boldsymbol{X}_D; Y_t) = I(\boldsymbol{X}_d; Y_t|\boldsymbol{X}_t\boldsymbol{X}_D),$$

where $I(\boldsymbol{X}_d; Y_t|\boldsymbol{X}_t\boldsymbol{X}_D)$ is the Conditional Mutual Information (*i.e.*, CMI) between $\boldsymbol{X}_d$ and $Y_t$ given $\boldsymbol{X}_t\boldsymbol{X}_D$ (Brown et al., 2012). Substituting the equation above into Eq. (2b), we define our information-theoretic data valuation metric, namely MShapley Conditional Mutual Information (*i.e.*, MShapley-CMI), as follows.

**Definition 5 (MShapley-CMI).** *Given a VFL task modeled by a master-managed cooperative prediction game, the MShapley-CMI value of data party $d$, which evaluates the contribution of the data party to the prediction of task label $Y_t$, is given by*

$$\varphi_d = \sum_{D \subseteq (\boldsymbol{\mathcal{D}} \setminus \{d\})} \frac{|D|!(|\boldsymbol{\mathcal{D}}| - |D| - 1)!}{|\boldsymbol{\mathcal{D}}|!} I(\boldsymbol{X}_d; Y_t | \boldsymbol{X}_t \boldsymbol{X}_D), \ \forall\, d \in \boldsymbol{\mathcal{D}}. \tag{6}$$

The MShapley-CMI metric given in Definition 5 is for evaluating the value of a data party. For the sake of completeness, by considering that the task party is always the first party entering the game, we can define the value of the task party as

$$\varphi_t = v(\{t\}) = H(Y_t) - H(Y_t|\boldsymbol{X}_t) = I(\boldsymbol{X}_t; Y_t), \tag{7}$$

where $I(\boldsymbol{X}_t; Y_t)$ is the mutual information between $\boldsymbol{X}_t$ and $Y_t$. Next, we analyze the properties of the proposed MShapley-CMI metric.

**Proposition 1.** *The MShapley-CMI metric has the following properties:*

- ***Master-aware Efficiency***: *The sum of the MShapley-CMI values across all data parties is equal to the mutual information between $\boldsymbol{X}_{\boldsymbol{\mathcal{D}}}\boldsymbol{X}_t$ and $Y_t$, subtracting the MShapley-CMI value of the task party, i.e.,*

$$\sum_{d \in \boldsymbol{\mathcal{D}}} \varphi_d = I(\boldsymbol{X}_{\boldsymbol{\mathcal{D}}} \boldsymbol{X}_t; Y_t) - \varphi_t. \tag{8}$$

- ***Symmetry***: *If data parties $d$ and $d'$ make the same contributions to the prediction of $Y_t$, conditioned on any coalition $S = \{t\} \cup D$, then they have the same MShapley-CMI value, i.e.,*

$$\textit{If } I(\boldsymbol{X}_d; Y_t | \boldsymbol{X}_D \boldsymbol{X}_t) = I(\boldsymbol{X}_{d'}; Y_t | \boldsymbol{X}_D \boldsymbol{X}_t), \forall D \subseteq (\boldsymbol{\mathcal{D}} \setminus \{d, d'\}), \ \textit{then } \varphi_d = \varphi_{d'}. \tag{9}$$

- ***Null-player**: If a data party makes no contribution to the prediction of $Y_t$, conditioned on any coalition $S = \{t\} \cup D$, then its MShapley-CMI value equals zero, i.e.,*

$$\textit{if } I(\boldsymbol{X}_d; Y_t | \boldsymbol{X}_D \boldsymbol{X}_t) = 0, \forall D \subseteq (\boldsymbol{\mathcal{D}} \setminus \{d\}), \textit{ then } \varphi_d = 0. \tag{10}$$

- ***Null-duplication**: Let $\boldsymbol{X}_d = (\boldsymbol{X}_d^1, \boldsymbol{X}_d^2)$ be the feature matrix of data party d. Suppose d duplicates $\boldsymbol{X}_d^2$ to create a forged feature matrix $\boldsymbol{X}_{\tilde{d}} = (\boldsymbol{X}_d^1, \boldsymbol{X}_d^2, \boldsymbol{X}_d^2)$. Let $\varphi_d$ and $\varphi_{\tilde{d}}$ denote the MShapley-CMI values calculated using the original feature matrix $\boldsymbol{X}_d$ and the forged matrix $\boldsymbol{X}_{\tilde{d}}$, respectively. We have*

$$\varphi_{\tilde{d}} = \varphi_d. \tag{11}$$

Proof. See Appendix B.

Proposition 1 shows several desirable properties of the MShapley-CMI metric for data valuation in a VFL task. Specifically, the master-aware efficiency property stipulates that only the contribution of additional features beyond those possessed by the task party will be allocated to data parties, thereby addressing the substitution effect between the task party and data parties. The symmetry property specifies that two data parties receive equal valuation if they make the same contribution to the prediction of $Y_t$. According to the null-player property, MShapley-CMI assigns zero value to a data party that does not offer any useful features for the prediction of $Y_t$. Finally, the null-duplication property property safeguards against the unfair acquisition of added value by a data party through malicious duplication of its features.

In sum, MShapley-CMI is a task-specific and model-free data valuation metric that takes into account of the substitution effect. It is task-specific because it evaluates the value of a data party based on the mutual information between the data party's feature matrix $\boldsymbol{X}_d$ and task label $Y_t$ (Eq. (6)). Moreover, the mutual information measure in Eq. (6) is independent of any machine learning model. Thus, MShapley-CMI is a model-free metric, unlike previous metrics that rely on the execution of a specific machine learning model for data valuation (Wang et al., 2019; Fan et al., 2022). Finally, as discussed before, MShapley-CMI considers the substitution effect because it evaluates a data party's marginal contribution to a prediction game, given the task party always joining the game before the data party. These merits make MShapley-CMI a promising data valuation metric, which is particularly important to fostering legitimate data markets, a trillion-dollar industry in the digital economy era (Chui et al., 2014).

## Federated Computation of MShapley-CMI

### Analysis of the Computation of MShapley-CMI

According to Definition 5, the challenge of computing the MShapley-CMI value lies in the computation of conditional mutual information $I(\boldsymbol{X}_d; Y_t|\boldsymbol{X}_D\boldsymbol{X}_t)$, which involves data distributed across task and data parties. By the definition of conditional mutual information, $I(\boldsymbol{X}_d; Y_t|\boldsymbol{X}_D\boldsymbol{X}_t)$ in Eq. (6) can be written as:

$$I(\boldsymbol{X}_d; Y_t|\boldsymbol{X}_D\boldsymbol{X}_t) = \sum_{\boldsymbol{x}_d,\boldsymbol{x}_D,\boldsymbol{x}_t,y_t} p(\boldsymbol{x}_d\boldsymbol{x}_D\boldsymbol{x}_t y_t) \log \frac{p(\boldsymbol{x}_d y_t|\boldsymbol{x}_t\boldsymbol{x}_D)}{p(\boldsymbol{x}_d|\boldsymbol{x}_t\boldsymbol{x}_D)p(y_t|\boldsymbol{x}_t\boldsymbol{x}_D)}, \tag{12}$$

where $D \subset \boldsymbol{\mathcal{D}}$ denotes a coalition of data parties, $\boldsymbol{x}_d$, $\boldsymbol{x}_D$, and $\boldsymbol{x}_t$ respectively represent values of features held by data party $d$, coalition of data parties $D$, and task party $t$, $y_t$ is a value of task label $Y_t$, and the summation is over all possible values of features held by $d$, $D$, and $t$, as well as all possible values of $Y_t$. In this equation, $p()$ denotes the probability of observing a combination of values. For example, $p(\boldsymbol{x}_d\boldsymbol{x}_D\boldsymbol{x}_t y_t)$ is the probability of observing a combination of $\boldsymbol{x}_d$, $\boldsymbol{x}_D$, $\boldsymbol{x}_t$, and $y_t$.

For values of discrete features, $\boldsymbol{x}$, the maximum likelihood estimation of the probability of observing $\boldsymbol{x}$ is $\hat{p}(\boldsymbol{x}) = N(\boldsymbol{x})/n$, where $N(\boldsymbol{x})$ is the number of data samples containing $\boldsymbol{x}$ and $n$ is the total number of data samples. Continuous features can be discretized, after which we can proceed with maximum likelihood estimation (Brown et al., 2012). Therefore, the conditional mutual information can be estimated as

$$\hat{I}(\boldsymbol{X}_d; Y_t|\boldsymbol{X}_D\boldsymbol{X}_t) = \sum_{\boldsymbol{x}_d,\boldsymbol{x}_D,\boldsymbol{x}_t,y_t} \hat{p}(\boldsymbol{x}_d\boldsymbol{x}_D\boldsymbol{x}_t y_t) \log \frac{\hat{p}(\boldsymbol{x}_d y_t|\boldsymbol{x}_t\boldsymbol{x}_D)}{\hat{p}(\boldsymbol{x}_d|\boldsymbol{x}_t\boldsymbol{x}_D)\hat{p}(y_t|\boldsymbol{x}_t\boldsymbol{x}_D)}$$

$$= \frac{1}{n} \sum_{\boldsymbol{x}_d,\boldsymbol{x}_D,\boldsymbol{x}_t,y_t} N(\boldsymbol{x}_d\boldsymbol{x}_D\boldsymbol{x}_t y_t) \log \frac{N(\boldsymbol{x}_d\boldsymbol{x}_D\boldsymbol{x}_t y_t) \sum_{\boldsymbol{x}'_d \in \mathcal{X}_d, y'_t \in \mathcal{Y}_t} N(\boldsymbol{x}'_d\boldsymbol{x}_D\boldsymbol{x}_t y'_t)}{\sum_{y'_t \in \mathcal{Y}_t} N(\boldsymbol{x}_d\boldsymbol{x}_D\boldsymbol{x}_t y'_t) \sum_{\boldsymbol{x}'_d \in \mathcal{X}_d} N(\boldsymbol{x}'_d\boldsymbol{x}_D\boldsymbol{x}_t y_t)}, \quad (13)$$

where $N(\cdot)$ denotes the number of data samples containing given feature values, $n$ is the total number of data samples, $\mathcal{Y}_t$ represents the set of all possible values of task label $Y_t$, and $\mathcal{X}_d$ denotes the set of all possible values for the features held by data party $d$. According to Eq. (13), to estimate the conditional mutual information, we need to compute $N(\boldsymbol{x}_d\boldsymbol{x}_D\boldsymbol{x}_t y_t)$, and we have

$$N(\boldsymbol{x}_d\boldsymbol{x}_D\boldsymbol{x}_t y_t) = \left| \bigcap_{d' \in D \cup \{d\}} \left\{ i | \boldsymbol{x}_{d'[i]} = \boldsymbol{x}_{d'} \right\} \bigcap \left\{ j | \langle \boldsymbol{x}_{t[j]}, y_{t[j]} \rangle = \langle \boldsymbol{x}_t, y_t \rangle \right\} \right|, \quad (14)$$

where $\boldsymbol{x}_{d'[i]}$ denotes the feature vector of the $i$-th data sample held by data party $d'$, and $\boldsymbol{x}_{t[j]}$ and $y_{t[j]}$ represent the feature vector and task label value of the $j$-th data sample owned by task party $t$, respectively. Eq. (14) computes $N(\boldsymbol{x}_d\boldsymbol{x}_D\boldsymbol{x}_t y_t)$ by finding the intersection among data samples owned by data and task parties that satisfy the respective equality condition specified in the equation and then returning the cardinality of the intersection. Therefore, the challenge of computing the MShapley-CMI value becomes how to find the intersection among data samples distributed across data and task parties while preserving the privacy of these data samples.

**Deficiency of Existing Solution**

*Secure multi-party computation* (MPC) is a framework that supports collaborative computations using data from multiple parties but reveals nothing other than the computational results to any of the parties (Goldwasser, 1997). Among existing MPC protocols, *Private Set Intersection* (PSI) is particularly suitable for the computation of the MShapley-CMI value because it computes the intersection of element sets distributed across different parties while allows each party to keep its data locally (Kamara et al., 2014). Nevertheless, computing the MShapley-CMI value using PSI could violate the privacy of participating parties, which we elaborate next.

Consider a case of computing the cardinality of the intersection between data samples owned by data party $d$ with their feature vectors being $\langle 1, 0 \rangle$ (i.e., $\boldsymbol{x}_d = \langle 1, 0 \rangle$) and data samples held by task party $t$ with their labels being 1 (i.e., $y_t = 1$). That is, we aim to compute $N(\boldsymbol{x}_d y_t)$, a simplified form of Eq. (14). To calculate $N(\boldsymbol{x}_d y_t)$, PSI first requires $d$ to encrypt IDs of its data samples with feature vectors being $\langle 1, 0 \rangle$ and $t$ to encrypt IDs of its data samples with labels being $1$, by using a prior established encryption agreement. Next, a third-party server collects the encrypted IDs from $d$ and $t$, computes the intersection of the encrypted IDs, and returns the computed intersection to $d$ and $t$. Finally, using the encryption agreement, $d$ and $t$ decrypt the encrypted IDs in the intersection. Consequently, party $d$ discovers that labels of data samples with their IDs in the intersection are $1$, which is private information owned by task party $t$. Similarly, $t$ learns that feature values of data samples with their IDs in the intersection are $\langle 1, 0 \rangle$; thus, private feature information owned by $d$ is disclosed to $t$.

One possible fix to the issue discussed above is to let a third-party server return the cardinality of the intersection, rather than the intersection, to $d$ and $t$. However, the third-party server could be *untrustworthy* and falsify results returned to $d$ and $t$. For instance, if the cardinality of data samples sent by task party $t$ to a third-party server is $n_t$, the server can easily deceive $t$ with any forged intersection cardinality $\hat{n}_s$ that is less than $n_t$. In practice, a federated machine learning method is required to be sufficiently reliable to discern falsified results from any *untrustworthy* server (IEEEStd., 2020). To this end, a state-of-the-art MPC protocol proposed by Le et al. (2019) can authenticate results provided by a third-party server, determining whether they have been falsified or not. However, this protocol can only work with *two* parties. The data valuation problem could easily involve more than two parties, e.g., one task party and multiple data parties. Therefore, we need to design a novel protocol that can securely and reliably compute the cardinality of the intersection among data samples distributed across multiple parties.

**Multi-party Secure Cardinality Computation Protocol**

We propose a *multi-party secure cardinality computation protocol*, which can verify the veracity of cardinality results returned by a third-party server. Like the PSI protocol, our proposed protocol encrypts IDs of data samples at task and data parties, and then sends encrypted IDs to a server to compute the cardinality of the intersection of these IDs. Different from the PSI protocol, our protocol has the capability to verify the veracity of the computed cardinality for VFL tasks that involve multiple parties. Such capability is realized through three schemes propsoed below.

**Definition 6 (ID Duplication Scheme).** *The scheme replicates the ID of each data sample owned by a party $q_{ID}$ times, where $q_{ID}$ is a random number agreed by all parties. For example, if $q_{ID}$ is set to 3, the scheme would generate three IDs, 'u011', 'u012, and 'u013', for a data sample with ID 'u01'. The convention for numbering duplicated IDs is also agreed by all parties. It is noted that the value of $q_{ID}$ is unknown to third-party servers.*

Following the scheme, each party (task or data party) duplicates IDs of its data samples $q_{\mathrm{ID}}$ times. It then encrypts these duplicated IDs and send them to a third-party server for the computation of cardinality. In principle, a veracious cardinality computed by the server should be divisible by $q_{\mathrm{ID}}$, as each data sample is associated with $q_{\mathrm{ID}}$ IDs. Furthermore, since the server is unaware of $q_{\mathrm{ID}}$, it is difficult for the server to forge a cardinality that can pass the divisibility check.

To further mitigate the risk of a server generating a falsified cardinality, we propose the following scheme.

**Definition 7 (Redundant Computation Scheme).** *Under this scheme, a server is requested to calculate a cardinality multiple times. In each iteration, a party (task or data party) follows the ID duplication scheme to replicate IDs of its data samples $q_{ID}$ times, and encrypts and sends them to the server for cardinality computation. Since $q_{ID}$ is a random number generated in each iteration, its value varies from one iteration to another.*

As a server is tasked with computing the same cardinality multiple times, a party should receive the same cardinality result in each iteration. In other words, if a party receives inconsistent cardinality results across iterations, it implies that the results have been falsified. Note that a server is unaware of $q_{\text{ID}}$, and the value of $q_{\text{ID}}$ changes from one iteration to another. Consequently, it is hard for an untrustyworhty server to pass the consistency check.

There remains a loophole in our protocol. An untrustworthy server could return zero cardinality in each iteration. Zero cardinality is divisible by $q_{\text{ID}}$ and consistent across iterations, thereby passing both the divisibility and consistency checks. To address this issue, we further enhance our protocol with the following scheme.

**Definition 8 (Data Augmentation Scheme).** *Under this scheme, each party produces $n_r$ artificial IDs, where $n_r$ is a random number agreed by all parties. The convention for numbering these artificial IDs is also agreed by all parties. Each party then encrypts these artificial IDs and sends the encrypted IDs to a server for cardinality computation.*

By executing the data augmentation scheme, all parties share a common set of $n_r$ artificial IDs. Therefore, this set of artificial IDs should be in the intersection, and the cardinality computed by a server should be at least $n_r$. In other words, a party can easily detect a server's malicious behavior if its computed cardinality is less than $n_r$.

**Federated Computation of MShapley-CMI**

Built on the proposed multi-party secure cardinality computation protocol, we develop a method to compute the MShapley-CMI value of each data party in a privacy-preserving manner. According to Definition 5 and Eq. 13, to calculate the MShapley-CMI value of a data party, we need to compute cardinality $N(\cdot)$ for each possible combination of $\boldsymbol{x}_d$, $\boldsymbol{x}_D$, $\boldsymbol{x}_t$, and $y_t$. To this end, we develop a secure and verifiable algorithm by following the multi-party secure cardinality computation protocol. As illustrated in Algorithm 1, at task or each data party, the algorithm encrypts IDs of its data samples that satisfy its corresponding condition specified in $N(\boldsymbol{x}_d\boldsymbol{x}_D\boldsymbol{x}_t y_t)$. The encrypted IDs are then sent to a third-party server to compute $N(\boldsymbol{x}_d\boldsymbol{x}_D\boldsymbol{x}_t y_t)$ in a secure way. Finally, the algorithm em-

**Algorithm 1:** Cardinality computation and verification

**Input :** A task party $t$, data parties $\boldsymbol{\mathcal{D}}$, and a cardinality computation job $N(\boldsymbol{x}_d\boldsymbol{x}_D\boldsymbol{x}_t y_t)$, where $d \in \boldsymbol{\mathcal{D}}$, $D \subset \boldsymbol{\mathcal{D}}$ and $d \notin D$;

**Output:** The value of $N(\boldsymbol{x}_d\boldsymbol{x}_D\boldsymbol{x}_t y_t)$;

Each party $e \in \{t \cup d \cup D\}$ prepares the IDs $\mathcal{I}_e$ of its data samples satisfying its corresponding condition specified in $N(\boldsymbol{x}_d\boldsymbol{x}_D\boldsymbol{x}_t y_t)$, where $\mathcal{I}_e = \{i | \boldsymbol{x}_{e[i]} = \boldsymbol{x}_e\}$ for $e \in \{d \cup D\}$ and $\mathcal{I}_e = \{i | \langle \boldsymbol{x}_{e[i]}, y_{e[i]} \rangle = \langle \boldsymbol{x}_e, y_e \rangle\}$ for $e = t$;

Following the Redundant Computation Scheme (Definition 7), task party $t$ generates a random number $\tau$ as the number of iterations and synchronizes $\tau$ with all data parties $\boldsymbol{\mathcal{D}}$;

**for** $j \in [1, \tau]$ **do**

  Task party $t$ generates a random number $q_{\text{ID}}$ as the number of ID duplications;

  Task party $t$ generates $n_r$ artificial IDs (denoted as $\mathcal{I}^r$) according to the Data Augmentation Scheme (Definition 8);

  Task party $t$ sends $q_{\text{ID}}$ and $\mathcal{I}^r$ to all data parties $\boldsymbol{\mathcal{D}}$;

  **for** $e \in \{t \cup d \cup D\}$ ***parallel*** **do**

    $e$ applies the ID Duplication Scheme (Definition 6) and replicates each ID in $\mathcal{I}_e$ for $q_{\text{ID}}$ times: DUP$(\mathcal{I}_e, q_{\text{ID}}) \to \mathcal{I}_e^d$;

    $\mathcal{I}_e^N = \mathcal{I}_e^d \cup \mathcal{I}^r$;

    $e$ encrypts the IDs in $\mathcal{I}_e^N$ (*i.e.*, Enc$(\mathcal{I}_e^N) \to \mathcal{I}_e^E$) and sends $\mathcal{I}_e^E$ to a third-party server.

  **end**

  The server computes the cardinality: $n_s = |(\bigcap_{e \in \{t \cup d \cup D\}} \mathcal{I}_e^E)|$ and returns $n_s$ to task party $t$;

  Task party $t$ applies **non-negativity verification**:

  **if** $n_s - n_r < 0$ **then**

    **return** *malicious behavior is detected*

  **end**

  Task party $t$ applies **divisibility verification**:

  **if** $(n_s - n_r)$ % $q_{ID}$ $! = 0$ **then**

    **return** *malicious behavior is detected*

  **end**

  Task party $t$ computes $N_j(\boldsymbol{x}_d\boldsymbol{x}_D\boldsymbol{x}_t y_t) = \frac{n_s - n_r}{q_{\text{ID}}}$;

  Task party $t$ applies **consistency verification**:

  **if** $j > 1$ *and* $N_j(\boldsymbol{x}_d\boldsymbol{x}_D\boldsymbol{x}_t y_t) \neq N_1(\boldsymbol{x}_d\boldsymbol{x}_D\boldsymbol{x}_t y_t)$ **then**

    **return** *malicious behavior is detected*

  **end**

**end**

$N(\boldsymbol{x}_d\boldsymbol{x}_D\boldsymbol{x}_t y_t) \leftarrow N_1(\boldsymbol{x}_d\boldsymbol{x}_D\boldsymbol{x}_t y_t)$;

**return** $N(\boldsymbol{x}_d\boldsymbol{x}_D\boldsymbol{x}_t y_t)$;

ploys the ID Duplication, Redundant Computation, and Data Augmentation Schemes to verify the trustworthiness of the computed $N(\boldsymbol{x}_d\boldsymbol{x}_D\boldsymbol{x}_t y_t)$. By invoking Algorithm 1 repeatedly, we can compute cardinality $N(\cdot)$ for each possible combination of $\boldsymbol{x}_d$, $\boldsymbol{x}_D$, $\boldsymbol{x}_t$, and $y_t$. Subsequently, we can estimate conditional mutual information $I(\boldsymbol{X}_d; Y_t | \boldsymbol{X}_D \boldsymbol{X}_t)$ by plugging these computed cardinalities into Eq. 13. With estimated $I(\boldsymbol{X}_d; Y_t | \boldsymbol{X}_D \boldsymbol{X}_t)$, the MShapley-CMI value of a data party can be computed according to Definition 5.

The proposed method for computing MShapley-CMI has the following security properties:

- **Preserving the Privacy of Raw Data**: As described in Algorithm 1, only encrypted IDs of data samples from task and data parties are sent to a third-party server. The actual data samples remain securely stored at their respective owning parties.
- **Security against Data Inference Attack**: According to Algorithm 1, the task party receives the cardinality of an intersection computed by a server, rather than the actual data sample IDs in the intersection. Consequently, the task party is unable to infer any features owned by data parties, thereby avoiding the data inference attack elaborated in the subsection of "Deficiency of Existing Solution".
- **Security against Falsified Result Attack**: Armed with the multi-party secure cardinality computation protocol, Algorithm 1 can effectively verify the trustworthiness of results returned by a server. Proposition 2 gives the upper bound of the probability that a malicious server can successfully deceive the task party during the process of computing an MShapley-CMI value.

**Proposition 2.** *Let $p_f$ be the probability that a malicious server can successfully deceive the task party during the process of computing the MShapley-CMI value of a data party. We have*

$$p_f \leq \min(q_{ID})^{-\tau\cdot|\mathcal{Y}_t||\mathcal{X}_t|\cdot\Pi_{d\in\boldsymbol{\mathcal{D}}}|\mathcal{X}_d|}, \tag{15}$$

*where $\tau$ is the number of iterations defined in the Redundant Computation Scheme, $\min(q_{ID})$ represents the minimum number of ID duplications across iterations, and $\mathcal{Y}_t$, $\mathcal{X}_t$, and $\mathcal{X}_d$ $(\forall d \in \boldsymbol{\mathcal{D}})$ denote the sets of all possible values of $Y_t$, $\boldsymbol{X}_t$, and $\boldsymbol{X}_d$, respectively.*

Proof. See Appendix C.

Consider a scenario involving only one data party and one task party. The data party owns two binary features, and we have $|\mathcal{X}_d| = 2^2 = 4$. The task party holds a binary task label and a binary feature; thus, we have $|\mathcal{Y}_t| = |\mathcal{X}_t| = 2$. If we set both $\min(q_{\text{ID}})$ and $\tau$ to be 2, by Proposition 2, $p_f$ is no larger than $2^{-2\cdot 2\cdot 2\cdot 4} = 2^{-32} \approx 10^{-10}$. Thus, even under a very simple scenario, the probability that a malicious server can successfully deceive the task party is slim. Real world scenarios are considerably more complex, involving numerous data parties, each with many more features. Consequently, in the real world, the probability that a malicious server can successfully deceive the task party is significantly less than $10^{-10}$. Proposition 2 demonstrates the strong security of our MShapley-CMI computation method against falsified result attacks by a malicious server. To the best of our knowledge, our study is among the first to design a verifiable federated method for multi-party secure cardinality computation against untrustworthy servers. Although the method is developed for the computation of MShapley-CMI values, it can also be applied to various other tasks involving secure and verifiable multi-party cardinality computation.

**Table 2. Statistics of experimental datasets**

| | *Wine* | *Parkinsons* | *Spect* | *Breast* | *Music* | *Credit* |
|---|---|---|---|---|---|---|
| *#features* | 13 | 22 | 22 | 30 | 57 | 23 |
| *#labels* | 3 | 2 | 2 | 2 | 10 | 2 |
| *#samples* | 178 | 195 | 267 | 569 | 1000 | 30000 |

## 5 EXPERIMENTS

We conduct extensive experiments to evaluate the effectiveness and computational efficiency of FedValue using six real-world datasets, including *Wine* (Aeberhard et al., 1994), *Parkinsons* (Little et al., 2007), *Spect* (Kurgan et al., 2001), *Breast* (Mangasarian et al., 1995), *Music*[5], and *Credit* (Yeh and Lien, 2009). Table 2 outlines the statistics of these datasets. Following Brown et al. (2012), we discretize continuous features into five equal-width categorical bins to facilitate MShapley-CMI computation. We also propose approximate computation techniques to improve the computational efficiency of FedValue for practical considerations.

[5]https://www.kaggle.com/andradaolteanu/gtzan-dataset-music-genre-classification

## Effectiveness of FedValue

In this section, we mainly evaluate the efficacy of the proposed MShapley-CMI metric in gauging the contribution of a data party to a prediction task.

### The Effectiveness of MShapley-CMI in Data Valuation

**Evaluation benchmarks**. State-of-the-art data valuation metrics, such as SHAP, require the training of a specific machine learning model, rendering them model-dependent. To the best of our knowledge, MShapley-CMI is the first model-free, task-specific data valuation metric. Since there is no prior model-free metric for comparison, we benchmark MShapley-CMI against existing model-dependent metrics. If data valuation results by MShapley-CMI are closely correlated with those obtained by state-of-the-art model-dependent metrics, it demonstrates the effectiveness of MShapley-CMI. This is because MShapley-CMI achieves performance comparable to state-of-the-art model-dependent metrics without relying on any particular machine learning model.

For benchmarks, we leverage SHAP (Lundberg and Lee, 2017) to value data parties given a specific prediction model, since it is the most widely used model-dependent metric. We implement various benchmarks by training multiple prediction models, including Support Vector Machine (SVM), Gradient Boosting Decision Tree (GBT), Logistic Regression (LR), Random Forest (RF), and Neural Networks (NN). Let $m$ denote one of the prediction models and SHAP-$m$ represent a SHAP-based data valuation metric dependent on model $m$. Following Fisher et al. (2019b), we also select a subset of well-performing models and average their SHAP values for a data party as another benchmark, denoted as SHAP-ensemble. A model is considered well-performing if its accuracy is within a threshold $\epsilon$ of the best model.

**Experimental setup**. We simulate a task party and multiple data parties to carry out experiments on each of the six datasets. Generally, a dataset's features are randomly selected and evenly distributed to all parties, while the label variable is assigned to the task party. The number of features per party is set to 1, 2, and 3 for robust validation. *Pearson correlation* is adopted to assess the cor-

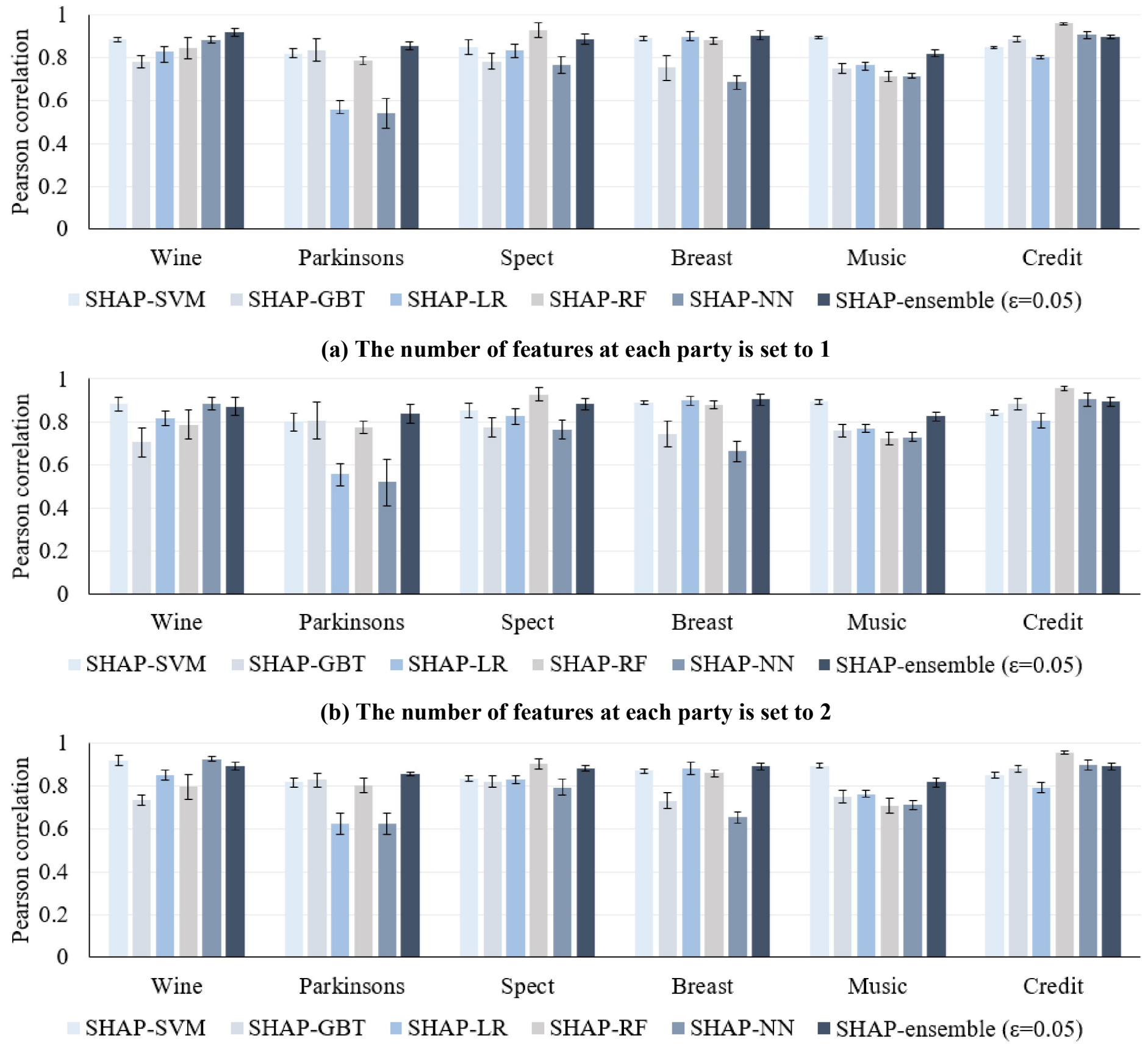

**(a) The number of features at each party is set to 1**

**(b) The number of features at each party is set to 2**

**(c) The number of features at each party is set to 3**

**Figure 4. Correlations between MShapley-CMI and model-dependent SHAP values.**

relation between data parties' MShapley-CMI values and SHAP-$m$ values. The threshold $\epsilon$ of SHAP-ensemble is defaulted to 0.05 and 0.02 for robustness check. Moreover, we randomly select 80% samples per dataset to run an experiment for each setting; by doing so, we can repeat the experiment 50 times with different data samples to obtain the mean and standard deviation of evaluation results for robustness check.

**Main results**. Fig. 4 displays the correlations between MShapley-CMI and SHAP values across various benchmarks for all six datasets. Across different datasets and settings of party feature numbers, the results reveal generally strong correlations between MShapley-CMI and various model-dependent SHAP benchmarks. Specifically, 91.7% and 65.7% of all the correlation values between MShapley-CMI and SHAP benchmarks surpass 0.7 and 0.8, respectively. Furthermore, the cor-relations between MShapley-CMI and SHAP-ensemble are almost consistently above 0.8 across datasets. These observations indicate that as a model-free metric, MShapley-CMI performs comparably to model-dependent metrics for data valuation in VFL tasks.

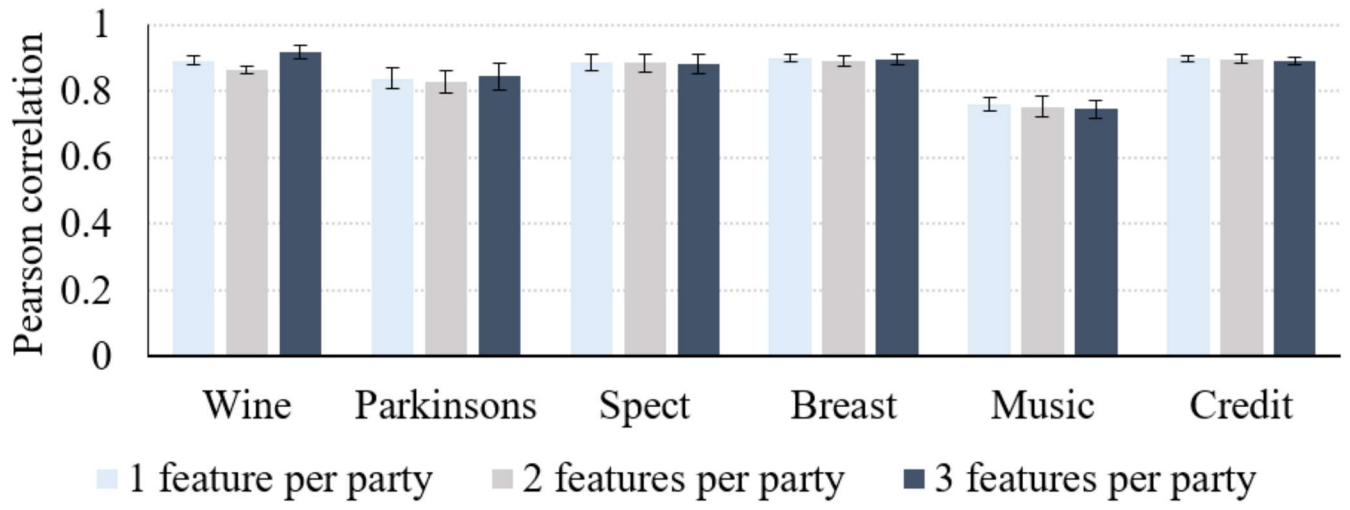

**Figure 5. Correlations between MShapley-CMI and SHAP-ensemble ($\epsilon = 0.02$).**

Interestingly, we discover that MShapley-CMI correlates more strongly with SHAP values from high-performance models (Table 8 in Appendix reports the model performance). For instance, on the Parkinsons dataset, the GBT model (0.908 accuracy) outperforms the NN model (0.856 accuracy). Correspondingly, MShapley-CMI exhibits a higher correlation with SHAP-GBT versus SHAP-NN. This finding also explains the close alignments between MShapley-CMI and SHAP-ensemble, since SHAP-ensemble averages well-performing models' party values. We also reset the SHAP-ensemble threshold $\epsilon$ to 0.02 across varying the number of features per party and find the consistent results shown in Fig. 5. Note that calculating SHAP requires collaboratively training actual models, which violates the model-free requirement for a data valuation in the initial stage of a VFL task. By avoiding model training while achieving similar performance, MShapley-CMI is better suited for early-phase data valuation in VFL tasks.

## Computational Efficiency of FedValue

Computational efficiency is often critical for the practical implementation of federated methods, as inefficient methods can demand huge computational resources, discourage party participation, and limit privacy protections. There are two key classes of parameters that influence the efficiency of FedValue. The first class relates to the MShapley-CMI computation, including the number of data parties, features, and data samples. The second involves parameters for verifying the trustworthiness of server results, such as ID duplications, redundant computations, and artificial IDs. We then vary these parameters to characterize the efficiencies achieved by FedValue.

**Table 3. Default parameter setting for computational efficiency evaluation**

| | *Data parties* | *Features per party* | *Real data samples* | *ID duplications* | *Redundant tasks* | *Artificial IDs* |
|---|---|---|---|---|---|---|
| # | 5 | 2 | 10,000 | 3 | 2 | 1-50 |

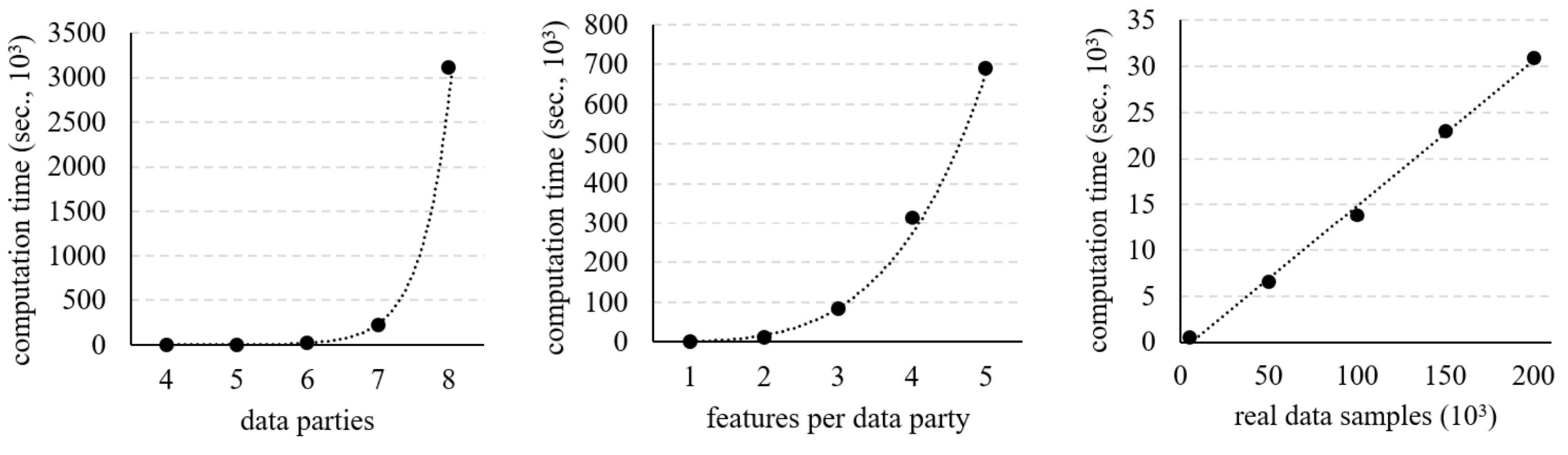

**Figure 6. Computation time by MShapley-CMI parameters.**

**Experimental setup**. To evaluate computation efficiency, we simulate a task party and multiple data parties using a setup similar to that used for effectiveness evaluation. Each party randomly selects a certain number of features, and we assume the sample IDs for these parties have been synchronized. Table 3 shows the default parameter settings. We vary one parameter at a time while fixing the others at their defaults, in order to assess how each parameter impacts the computation efficiency of FedValue. The experiments are carried out on an ordinary laptop with AMD Ryzen R7 4800HS (2.9GHz) and 16GB RAM.

**Main results**. Fig. 6 shows the computation time of FedValue concerning the three key parameters related to MShapley-CMI computation. Specifically, Fig. 6a and Fig. 6b illustrate that the computation time increases exponentially with the number of data parties and the number of features per party, respectively. Fig. 6c displays a linear correlation between FedValue's running time and the number of data samples. For instance, when each party contains 1,000,000 samples, FedValue can complete the data valuation in 4 hours with the other parameters set at default values. Additionally, Fig. 7 shows that FedValue exhibits linear scaling regarding the number of ID duplications, redun-dant computations, and artificial IDs – all three verification parameters.

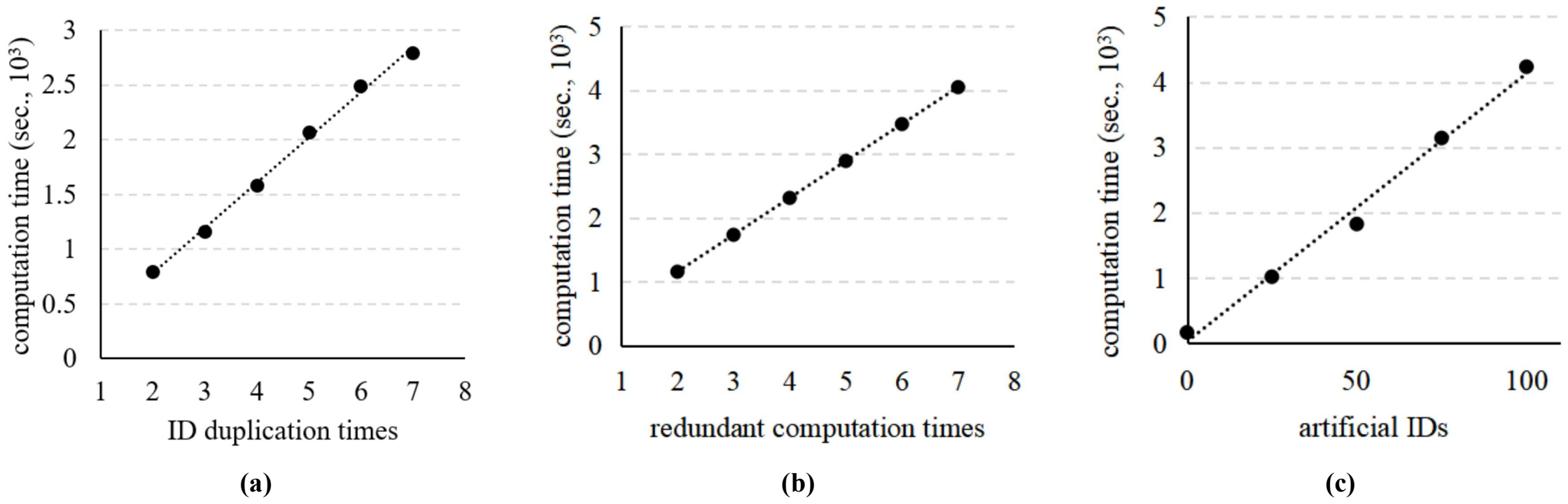

**Figure 7. Computation time by privacy-preserving related parameters.**

While linear time complexity usually indicates favorable scalability, exponential scalability may require acceleration techniques for practical applications (Cormen et al., 2022). Our experiments demonstrate that FedValue can maintain its efficiency even when parties have massive data samples. FedValue also exhibits promising efficiency in enabling secure and verifiable federated computations. Nevertheless, as the number of data parties or features per party grows substantially, FedValue may face computational efficiency challenges. Therefore, in the next subsection, we propose approaches to mitigate potential computational efficiency issues with FedValue.

### Efficiency Improvement by Approximation Techniques

We employ two approximation techniques, *i.e.*, *Shapley sampling* and *feature dimension reduction*, to accelerate FedValue for practical data valuation scenarios with many data parties or features at each party.

Inspired by prior work on approximating Shapley values (Strumbelj and Kononenko, 2013), we leverage the Shapley sampling method to sample and enumerate a subset of permutations of data parties. A data party $d$'s value $\hat{\varphi}_d$ is then estimated by averaging the conditional mutual information across the set of sampled data party permutations. That is,

$$\hat{\varphi}_d = \frac{1}{|\mathcal{O}|} \sum_{o \in \mathcal{O}} I(\boldsymbol{X}_d; Y_t | \boldsymbol{X}_{D_o} \boldsymbol{X}_t), \tag{16}$$

where $\mathcal{O}$ is a sampled set of permutations of data parties, $o$ is a specific permutation of data parties in $\mathcal{O}$, and $\boldsymbol{X}_{D_o}$ is the set of data parties that precede the party $d$ in $o$.

Dimension reduction techniques such as principal component analysis (PCA) (Martínez and Kak, 2001) can also improve computation efficiency (Fodor, 2002). In our case, before calculating MShapley-CMI, parties with high-dimensional features can independently run PCA to acquire a low-dimensional feature vector of principal components. FedValue can then be accelerated by operating on this low-dimensional feature vector for data valuation in VFL.

**Experimental setup**. We denote the version of FedValue using approximate computation techniques as fast-FedValue and validate these approximate techniques by comparing fast-FedValue to the original FedValue. Ideally, fast-FedValue should achieve similar data valuation results to FedValue but with improved efficiency. We use Mean Absolute Percentage Error (MAPE) to quantify differences in data valuation between fast-FedValue and FedValue. The datasets *Breast* and *Music*, which have the highest feature dimensions, are used for empirical validation.

**Main result**. In general, Fast-FedValue with Shapley sampling cannot attain the precise value of MShapley-CMI if the number of sampled perturbations is small. Therefore, we focus on how many samplings are needed for fast-FedValue to achieve an acceptably low MAPE. Fig. 8a displays how the MAPE changes with the number of samplings. As it shows, the MAPE drops sharply when the number of samplings increases from 100 to 1200, then steadily converges to a small value with further increases. Fig. 8b illustrates the linear relationship between the number of samplings and MShapley-CMI computation time. Therefore, when there are excessive data parties, we can apply Shapley sampling and set a reasonable number of samples for MShapley-CMI computation to achieve comparable valuation effectiveness within acceptable computation time.

In addition, we construct fast-FedValue using PCA to reduce the feature dimensions of the data parties. Fig. 9a displays that the MAPE between FedValue and fast-FedValue decreases rapidly as the number of retained PCA components increases from 2 to 4. This indicates that fast-FedValue can better approximate the true MShapley-CMI when more informative components are kept. Fig. 9b

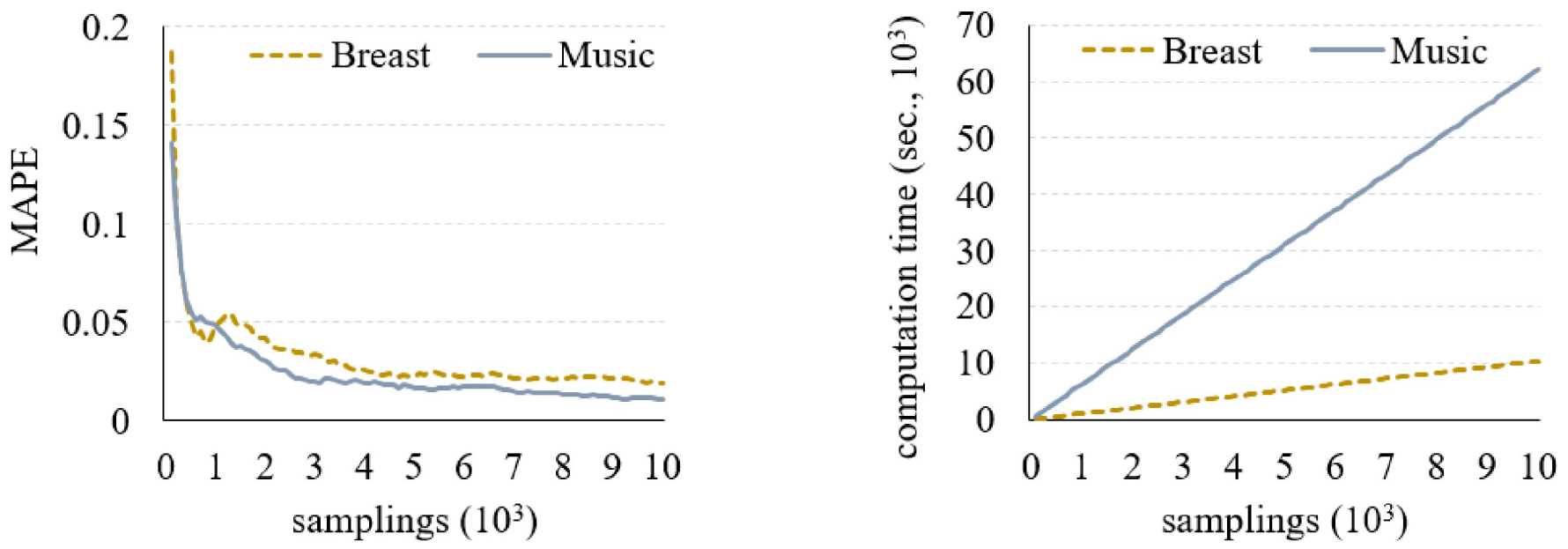

**(a) Data valuation effectiveness** **(b) Computational efficiency**

**Figure 8. FedValue acceleration performance by using Shapley sampling.**

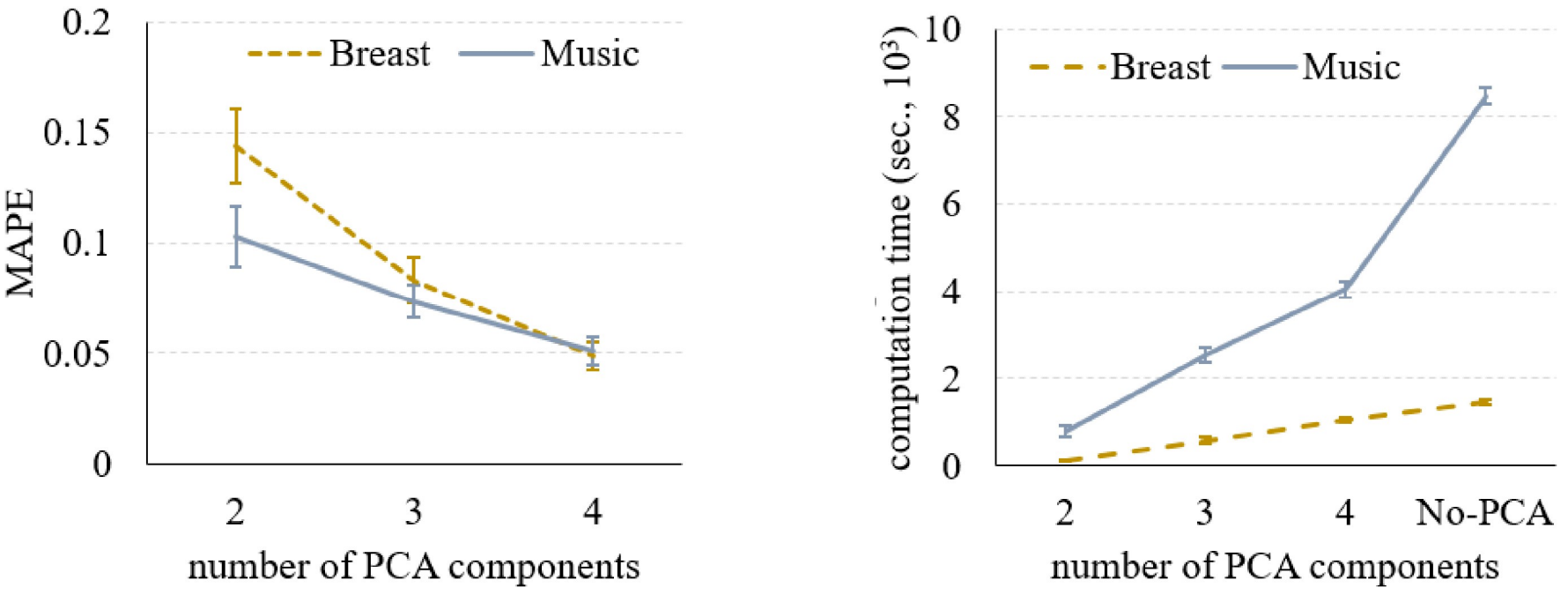

**(a) Data valuation effectiveness** **(b) Computational efficiency**

**Figure 9. FedValue acceleration performance by using PCA.**

compares fast-FedValue and FedValue by the computation time and demonstrates an exponential decrease of computation time for FedValue with fewer PCA components per party. In brief, for data parties with high-dimensional features, FedValue can balance effectiveness and efficiency by having them apply PCA and retain an appropriate number of principal components.

## 6 FEDERATED RECOMMENDATION: A CASE STUDY OF FEDVALUE

In this section, we conduct a case study to demonstrate the practical value of FedValue for data party selection in a vertical federated recommendation task. Specifically, we aim to answer three key questions: 1) Can FedValue help to select suitable data parties for effective federated recommendations given a certain budget? 2) Can FedValue address the substitution effect appropriately? 3) Can FedValue discern and exclude free-rider parties that lack useful data?

## Experimental Settings

### Overview of the Federated Recommendation Task

We assume a movie agency has collected demographic data on all members and gathered preferences of a small subset of members for a group of upcoming movies through marketing promotions. The agency aims to identify target customers[6] among its broader members for bundled movie recommendations. However, the agency's limited internal data may be inadequate to identify target customers accurately. Hence, the agency plans to launch a vertical federated recommendation task and incorporate some external data sources given a certain amount of budget to augment the recommendations.

Multiple data holders (*e.g.*, MovieLens, YouTube, and IMDB) contain rich movie preference data about the agency's members and may wish to monetize their data by engaging in the federated recommendation task. As such, the agency requires an effective strategy for selecting data holders that can maximize recommendation improvements within budget constraints. Without loss of generality, we formulate the federated recommendation task as a federated binary classification problem, where the goal is to identify target customers for the recommended movies. The Gradient Boosted Trees (GBT) model is employed for classification since it performs generally well in our experiments.

### Dataset and Party Configuration

We leverage a movie rating benchmark dataset, *i.e.*, *MovieLens-1M* (Harper and Konstan, 2015), to conduct experiments. This dataset contains around 1 million ratings across 4,000 movies provided by 6,000 users. It also includes demographic information for each user such as gender and age. Based on this dataset, we configure the federated learning setup as:

**Task party**. Consider a movie agency serving as the task party in seeking a high-performance recommendation model for a bundle of new movies. We randomly sample a certain number of movies ($N = 20$ by default) from the MovieLens-1M dataset as the new movies to recommend. Users are labeled as "target" customers if they previously rated at least one of these movies as a 4 or 5, other-wise as “non-target”. We also sample a certain proportion ($r_t = 20\%$ by default) of users to construct a training set with labeled target and non-target customers. These users represent engaged customers who previewed and provided feedback on the movies through initial marketing promotions. The remaining users are assigned to a test set, simulating the broader pool of potential customers upon official release. Acting as the task party, the agency aims to train a binary federated classification model using all users’ demographic data, label information from the training set, and supplemental user features from the selected data parties. The trained model is expected to classify the test users into target or non-target customers for precision marketing of the new movies.

[6]Members who express no interest in any of the upcoming films are considered *non-target customers*, otherwise *target customer*.

**Data parties**. Data holders who can share historical user-movie ratings are considered candidate data parties. To configure a data party, we randomly select a subset of movies (excluding those already selected as new movies) from the MovieLens-1M dataset and leverage the user-movie rating data as user preference features at the party. All data parties and the task party share an identical set of users. We configure half of the data parties as *low-volume parties* encompassing a smaller subset of movies (1% of the total by default), and the remaining half as *high-volume parties* with relatively more movies (10% by default). Consequently, every data party holds a user-movie rating matrix $\mathbf{M}_d$, where each element falls within the 0-5 range (0 means no rating). Due to the high-dimensional and sparse nature of $\mathbf{M}_d$, every data party applies PCA and employs the top-$K$ ($K = 2$ by default) components to represent users’ features.

**Incentive Schemes**

We consider two monetary incentive schemes commonly applied in an FL task.

**Fixed incentives (FIX)**. In this scheme, the task party $t$ pays a fixed amount of monetary incentive $\alpha$ to every selected data party $d$. Despite being simple, FIX is widely used in realistic collaborator recruitment processes, *e.g.*, crowdsourcing platforms like Amazon Mechanical Turk[7].

**Varied incentives (VAR)**. The VAR scheme allows varying incentives for different data parties. Referring to Zhan et al. (2020), the incentives for low-volume and high-volume data parties are set to

[7] https://docs.aws.amazon.com/AWSMechTurk/latest/AWSMechanicalTurkRequester/mturk-hits-attributes.html

$\alpha$ and $2\alpha$, respectively, in our experiments.

**Party Selection Strategies**

Suppose that the task party $t$ has a limited budget $B$ set at $k \cdot \alpha$ ($k = 2$ by default), and all the candidate data parties in $\mathcal{D}$ consent to the adopted incentive scheme. We explore four diverse strategies for iteratively selecting data parties from a basket of candidates using different selection metrics until the budget is used up. Specifically, FedValue and VOLUME strategies prioritize the parties with larger MShapley-CMI values and data volume, respectively. RANDOM strategy greedily chooses affordable data parties at random. VF-PS (Jiang et al., 2022), a cutting-edge party selection strategy, relies on mutual information. VF-PS randomly creates numerous party groups, averaging the mutual information between the task label and the collected features from all parties within each group to score the parties; VF-PS takes the mean score of a party across different groups for party selection. We emphasize that model-dependent methods (*e.g.*, SHAP) are not viable for scenarios where party valuation or selection precedes modeling, making them incomparable.

**Experimental Scenarios and Comparison Metrics**

Three experimental scenarios have been developed to address the initial three questions in this section. Firstly, an ideal scenario is established wherein all data party candidates hold valuable data distinct from that of the task party. To assess FedValue's ability to handle the substitution effect, we develop another scenario involving some redundant parties whose data have been possessed by the task party. We expect that FedValue can refrain from selecting these redundant parties. We setup the third scenario by introducing free-rider parties that attempt to deceive the task party by providing non-valuable data (*i.e.*, fabricated use-movie ratings) (Fraboni et al., 2021; Kairouz et al., 2021). We vary the ratios of redundant and free-rider parties among all data party candidates (40% by default) in the second and third scenarios, respectively.

We use *accuracy* and *AUC* to assess the recommendation performance of different party selection strategies in all three scenarios. Additionally, we introduce the *redundant party selection probability* (i.e., RPSP) and the *free-rider selection probability* (i.e., FRSP) to evaluate the effectiveness of party selection strategies in excluding redundant and free-rider parties, respectively. RPSP/FRSP calculates the mean proportion of the redundant/free-rider party among all the selected parties. In essence, a higher RPSP or FRSP value signifies a less effective party selection method for selecting less valuable data parties given the same budget.

**Table 4. Parameter Settings for Federated Recommendation**

| Variable | budget | #movies for recommend. | #data parties | #features per data party | low ratio of movies | high ratio of movies | redundant party ratio | free-rider party ratio |
|---|---|---|---|---|---|---|---|---|
| Default value | $2\alpha$ | 20 | 5 | 2 | 1% | 10% | 20% | 20% |
| Varying value | $\alpha$,$3\alpha$ | 30,40 | 4,6 | 1,3 | – | 5%,20% | 40%,60% | 40%,60% |

Finally, we execute federated recommendations by employing parties determined by various party selection strategies to compare their efficacy across three experimental scenarios. We repeat each experiment 20 times by randomly sampling data for party configurations and report the average result with standard deviations. Detailed parameter settings are outlined in Table 4. Throughout the main texts, we vary the budget of the task party from $\alpha$ to $3\alpha$ under both FIX and VAR incentive schemes, while fixing other parameters at default values. Robustness checks examining variations in other parameters are presented in the Appendix.

## Results

Table 5 reports the performance of different party selection strategies in federated recommendations within the first scenario, where all data party candidates hold useful data for the task party. The results show that FedValue consistently achieves significantly higher accuracy and AUC values compared with other party selection strategies across different budget settings under both FIX and VAR incentive schemes. For example, with $2\alpha$ budgets, FedValue achieves an accuracy of 0.748 and 0.750 under FIX and VAR schemes, respectively, slightly surpassing VF-PS's accuracy of 0.748 and 0.749 using $3\alpha$ budgets, which means lower investments through FedValue can yield higher returns.

The results also suggests the special advantage of FedValue with tight budgets. For example, the AUC improvements of FedValue over the best benchmark VF-PS under the FIX scheme diminish successively from 0.01 to 0.002 as the budget expands from $\alpha$ to $3\alpha$. Since the FIX scheme allocates an equal incentive to all selected parties, a more limited budget leads to the selection of fewer data parties. Therefore, the results highlight FedValue's ability to prioritize the selection of the most valuable data parties.

**Table 5. Results of federated recommendation without free-rider or redundant candidate data parties. (The significance of $t$-test is for the comparison between FedValue and the second-best benchmark, $^{*}$ < 0.1, $^{**}$ < 0.05, $^{***}$ <0.01).**

| *incentive scheme* | *budget* | *accuracy* $\alpha$ | *accuracy* $2\alpha$ | *accuracy* $3\alpha$ | *AUC* $\alpha$ | *AUC* $2\alpha$ | *AUC* $3\alpha$ |
|---|---|---|---|---|---|---|---|
| **FIX** | **RANDOM** | 0.728 (±0.030) | 0.741 (±0.025) | 0.746 (±0.022) | 0.721 (±0.019) | 0.752 (±0.017) | 0.766 (±0.017) |
| | **VOLUME** | 0.737 (±0.031) | 0.741 (±0.031) | 0.746 (±0.025) | 0.726 (±0.015) | 0.739 (±0.020) | 0.761 (±0.018) |
| | **VF-PS** | 0.734 (±0.030) | 0.744 (±0.025) | 0.748 (±0.023) | 0.731 (±0.029) | 0.754 (±0.025) | 0.766 (±0.021) |
| | **FedValue** | **0.744**$^{***}$ **(±0.026)** | **0.748**$^{**}$ **(±0.022)** | **0.750**$^{**}$ **(±0.022)** | **0.739**$^{**}$ **(±0.020)** | **0.760**$^{**}$ **(±0.021)** | **0.771**$^{***}$ **(±0.020)** |
| **VAR** | **RANDOM** | 0.723 (±0.031) | 0.736 (±0.027) | 0.743 (±0.023) | 0.719 (±0.026) | 0.740 (±0.017) | 0.758 (±0.018) |
| | **VOLUME** | 0.723 (±0.032) | 0.737 (±0.031) | 0.744 (±0.024) | 0.719 (±0.026) | 0.726 (±0.015) | 0.754 (±0.017) |
| | **VF-PS** | 0.728 (±0.029) | 0.738 (±0.026) | 0.743 (±0.023) | 0.730 (±0.030) | 0.743 (±0.025) | 0.760 (±0.020) |
| | **FedValue** | **0.734**$^{**}$ **(±0.027)** | **0.746**$^{***}$ **(±0.024)** | **0.749**$^{***}$ **(±0.021)** | **0.740**$^{***}$ **(±0.028)** | **0.744** **(±0.026)** | **0.766**$^{***}$ **(±0.019)** |

Table 6 presents the federated recommendation performance in the scenario with 20% redundant data party candidates. The results again reveal that FedValue consistently outperforms the other party selection strategies. Most importantly, FedValue is the only strategy that excludes *ALL* redundant data party candidates and achieves a redundant party selection probability (RPSP) of 0% across different experimental settings. It is also noteworthy that FedValue seems to gain more advantages over the second best strategy VF-PS in this scenario. Specifically, under the FIX scheme with a budget $\alpha$, the accuracy improvement of FedValue over VF-PS in this scenario ($\frac{0.730-0.714}{0.714} = 2.24\%$) exceeds that ($\frac{0.744-0.734}{0.734} = 1.36\%$) achieved in the scenario without redundant candidates by 65%. These re-sults verify the ability of FedValue to manage the substitution effects and rule out the invaluable data parties, thereby leading to a performance advantage.

**Table 6. Results of federated recommendation with 20% redundant data parties whose features are also held by the task party. (The significance of $t$-test is for the comparison between FedValue and the second-best benchmark, $^{*}<0.1$, $^{**}<0.05$, $^{***}<0.01$; RPSP stands for the redundant party selection probability.)**

| *incentive scheme* | *budget* | *accuracy* $\alpha$ | *accuracy* $2\alpha$ | *accuracy* $3\alpha$ | *AUC* $\alpha$ | *AUC* $2\alpha$ | *AUC* $3\alpha$ | *RPSP* $\alpha$ | *RPSP* $2\alpha$ | *RPSP* $3\alpha$ |
|---|---|---|---|---|---|---|---|---|---|---|
| **FIX** | **RANDOM** | 0.702 (±0.027) | 0.724 (±0.022) | 0.732 (±0.020) | 0.714 (±0.020) | 0.752 (±0.017) | 0.770 (±0.018) | 18% | 21% | 20% |
| | **VOLUME** | 0.707 (±0.034) | 0.724 (±0.026) | 0.732 (±0.022) | 0.709 (±0.025) | 0.736 (±0.021) | 0.764 (±0.019) | 20% | 20% | 20% |
| | **VF-PS** | 0.714 (±0.033) | 0.729 (±0.024) | 0.734 (±0.022) | 0.729 (±0.031) | 0.761 (±0.024) | 0.775 (±0.022) | 10% | 14% | 13% |
| | **FedValue** | **0.730**$^{***}$ (**±0.022**) | **0.735**$^{***}$ (**±0.022**) | **0.738**$^{**}$ (**±0.022**) | **0.746**$^{***}$ (**±0.022**) | **0.765** (**±0.020**) | **0.777** (**±0.019**) | **0%** | **0%** | **0%** |
| **VAR** | **RANDOM** | 0.699 (±0.029) | 0.715 (±0.024) | 0.727 (±0.022) | 0.715 (±0.025) | 0.738 (±0.017) | 0.759 (±0.017) | 19% | 20% | 20% |
| | **VOLUME** | 0.700 (±0.029) | 0.706 (±0.034) | 0.726 (±0.023) | 0.716 (±0.024) | 0.710 (±0.025) | 0.752 (±0.017) | 21% | 20% | 20% |
| | **VF-PS** | 0.711 (±0.037) | 0.717 (±0.033) | 0.730 (±0.024) | 0.730 (±0.044) | 0.741 (±0.036) | 0.766 (±0.024) | 15% | 10% | 15% |
| | **FedValue** | **0.725**$^{***}$ (**±0.022**) | **0.731**$^{***}$ (**±0.022**) | **0.736**$^{***}$ (**±0.022**) | **0.749**$^{***}$ (**±0.022**) | **0.755**$^{***}$ (**±0.025**) | **0.770**$^{*}$ (**±0.019**) | **0%** | **0%** | **0%** |

Table 7 presents the results when there exist 20% free riders among the candidate data parties. The results reveal that RANDOM and VOLUME fail to distinguish free riders from benign candidates, yielding an FRSP approximating the free-rider candidate ratio (20%). VF-PS achieves a relatively low FRSP below 5% but yet selects some free riders occasionally. Comparatively, FedValue obtains a free-rider selection probability (FRSP) of 0% across all settings, indicating its ability to reject free riders and resulting in the improvement of recommendation performance. For instance, under the VAR scheme with a budget of $\alpha$, FedValue's AUC surpasses RANDOM, VOLUME and VF-PS by 8.2%, 8.7% and 1.9%, respectively. These findings underscore the practical value of FedValue in safeguarding against free-rider candidates, enhancing model performances, and conserving limited budgets.

**Table 7. Results of federated recommendation with free-rider candidate data parties of 20%. (The significance of $t$-test is for the comparison between FedValue and the second-best benchmark, $^{*}$ < 0.1, $^{**}$ < 0.05, $^{***}$ <0.01; FRSP represents the free-rider selection probability.)**

| *incentive scheme* | *budget* | *accuracy* | | | *AUC* | | | *FRSP* | | |
|---|---|---|---|---|---|---|---|---|---|---|
| | | $\alpha$ | $2\alpha$ | $3\alpha$ | $\alpha$ | $2\alpha$ | $3\alpha$ | $\alpha$ | $2\alpha$ | $3\alpha$ |
| **FIX** | **RANDOM** | 0.707 (±0.039) | 0.737 (±0.028) | 0.744 (±0.025) | 0.681 (±0.020) | 0.739 (±0.019) | 0.757 (±0.018) | 19% | 20% | 20% |
| | **VOLUME** | 0.706 (±0.058) | 0.743 (±0.030) | 0.746 (±0.025) | 0.672 (±0.060) | 0.733 (±0.021) | 0.755 (±0.018) | 25% | 25% | 22% |
| | **VF-PS** | 0.735 (±0.031) | 0.743 (±0.027) | 0.748 (±0.023) | 0.730 (±0.023) | 0.755 (±0.024) | 0.768 (±0.018) | 0% | 4% | 3% |
| | **FedValue** | **0.740**$^{***}$ (±**0.029**) | **0.746**$^{**}$ (±**0.024**) | **0.749**$^{**}$ (±**0.023**) | **0.735**$^{**}$ (±**0.016**) | **0.761**$^{**}$ (±**0.019**) | **0.774**$^{***}$ (±**0.018**) | **0%** | **0%** | **0%** |
| **VAR** | **RANDOM** | 0.708 (±0.038) | 0.722 (±0.037) | 0.737 (±0.028) | 0.683 (±0.044) | 0.712 (±0.026) | 0.744 (±0.020) | 17% | 18% | 20% |
| | **VOLUME** | 0.709 (±0.038) | 0.707 (±0.057) | 0.739 (±0.028) | 0.680 (±0.044) | 0.672 (±0.059) | 0.741 (±0.021) | 18% | 25% | 20% |
| | **VF-PS** | 0.728 (±0.031) | 0.737 (±0.031) | 0.744 (±0.025) | 0.725 (±0.046) | 0.739 (±0.026) | 0.759 (±0.023) | 3% | 4% | 5% |
| | **FedValue** | **0.733**$^{**}$ (±**0.027**) | **0.742**$^{***}$ (±**0.029**) | **0.746**$^{*}$ (±**0.024**) | **0.739**$^{**}$ (±**0.021**) | **0.743** (±**0.021**) | **0.764**$^{**}$ (±**0.016**) | **0%** | **0%** | **0%** |

In summary, the case study verifies that FedValue can significantly enhance federated recommendation systems by effectively selecting appropriate data parties across various practical settings. As reported by Wang et al. (2022), even a 0.1% improvement in recommendation performance means a substantial increase in practical click-through rates for real-world businesses. Therefore, a federated recommendation system integrated with FedValue for party selection might yield significant monetary gains especially when considering that the global film industry (including box office and home entertainment revenue) has exceeded \$136 billion since 2018[8].

## 7 CONCLUSION

Federated learning is a burgeoning privacy-preserving paradigm for collaborative predictive analytics, garnering substantial attention from both industries and academia. Different from most existing studies on developing effective learning algorithms, we focus on addressing the pivotal issue of val-uating data in the initial stage of VFL tasks from a business perspective. This issue holds critical significance for the subsequent party selections and incentive mechanisms design but is unfortunately under-investigated. We emphasize the difficulty of data valuation in real-life businesses, as a task party lacks access to data from other parties yet needs crucial data value information to select parties before collaboratively training a concrete model with these parties. To overcome the challenge, we propose FedValue, a model-free but task-specific data valuation framework for VFL. We demonstrate that FedValue is a useful business decision support tool for a task party to choose appropriate data parties for high-performance VFL.

[8]https://en.wikipedia.org/wiki/Film_industry

Our study presents a few theoretical contributions. First, we scrutinize the issue of modeling a VFL task with the conventional cooperative game, which overlooks the substitution effect between candidate data parties and task parties. To address this issue, we propose a master-managed cooperative prediction game (MMCPG) to model a VFL task, considering the master role of a task party in a VFL. MMCPG could establish an effective theoretical foundation for future federated learning studies. Moreover, while the Shapley value allocation framework averages all players' marginal values across all possible orders in player permutations, it fails to account for the fact that the task party initiates a VFL task and only pays a data party for additional data that is useful for a predictive task. We extend the Shapley value and propose a valuation framework named MShapley tailored for VFL within MMCPG. MShapley prioritizes the task party in contribution allocation and averages the marginal values of data parties concerning all possible permutations that data parties follow the task party. MMCPG and MShapley are believed to enhance the modeling of broader multi-party cooperative activities. Lastly, given the crucial role of information/data value within information systems, our work has the potential to enrich the IS literature by analyzing and quantifying the value of data within a privacy-preserving collaborative paradigm for predictive analytics.

Our study also makes some methodological contributions. Primarily, we propose a novel practical FedValue framework to value data at each party of a VFL task in a privacy-preserving manner. In particular, rooted in information theory, we develop MShapley-CMI, a model-free but task-specific metric. In other words, MShapley-CMI can value data parties in a specific VFL task without training a model as accurately as those model-dependent metrics. Besides, inspired by the existing PSI computation protocols, we design a federated mechanism to compute MShapley-CMI in a privacy-preserving, secure, and verifiable way. This design keeps parties from data leakage and deception during computations even with untrustworthy servers. Moreover, we notice the computationally complex problem of MShapley-CMI and suggest using Shapley sampling and dimension reduction as boosters. The effectiveness and efficiency of FedValue are verified comprehensively by extensive experiments on six real-world datasets.

We analyze the practical value of FedValue for real-life businesses. To this end, we conduct a simulated case study of federated movie recommendations with various economic considerations including budgets, incentive mechanisms, and free-rider candidates. The case study shows that a movie agency can use FedValue to proficiently identify valuable data parties for building a federated recommendation system. Concretely, FedValue ranks high-quality data parties at the top of the selected list and is capable of excluding free-rider candidates, which is particularly important when the budget is tight. Given the report that a 0.1% level of recommendation improvement means a substantial raise of the practical click-through rates in real-life businesses (Wang et al., 2022), the monetary gain by FedValue should be significant given the over $100 billion global film market. Our work also sheds light on some general managerial implications for data valuation in VFL tasks. For example, we observe that party selection according to data volume is often ineffective and sometimes is even less competent than random selection. This highlights the intricate interplay between data value and specific predictive tasks.

The value of our method goes beyond federated movie recommendation. For example, in the content consumption market where a few super-giants like TouTiao, TicTok, and YouTube occupy the leading positions, a federated recommendation can help the companies improve their target marketing abilities in user growth and Ads placement by integrating more user behavioral data from electronic business companies like JingDong and telecoms like China Mobile. For another, the financial robo-advisor market is still a blue ocean in China, where federated recommendation can play a key role in finding high-value customers for banks' wealth management business and in promoting new financial products for mutual or private funds companies, with the auxiliary data for federated learning again coming from some online platforms or telecoms. In these applications, a data valuation mechanism like FedValue is in great need of designing incentive mechanisms to absorb high-quality data providers and for making optimal party selections before launching collaborative model training. We can further expand the application scenario of FedValue to more predictive analytics tasks where vertical federated learning plays an essential role, *e.g.*, systemic risk management across different financial markets, urban intelligent management across different governance sectors, COVID-19 infectious prediction based on multi-source data (onset, mobility, weather, *etc.*), among others. According to the Gartner's Hype Cycle for Privacy[9], federated learning is still on the rise to the peak and is expected to penetrate privacy-preserving predictive analytics in a variety of domains rapidly. Thus, FedValue has the potential to generate considerable business value in the future if practically adopted.

Our study has limitations and there are research issues merit future attention and efforts. Firstly, our work does not investigate the important incentive mechanisms issue for a VFL task, although data valuation is the key component in mechanism design. In the near future, we plan to take a step forward to this point based on the FedValue method and study the interactions between data valuation and optimal mechanism design, which in turn can improve FedValue greatly. Secondly, while we have proposed some practical acceleration methods for FedValue, the very nature of MShapley-CMI will incur big computational costs with the rise of the number of features and data parties. Therefore, we would like to explore whether there is more room for designing approximate calculation methods for MShapley-CMI or reducing ineffective computations for party coalitions with little impact on valuation precision. Finally, our work only provides a prototype design for VFL data valuation and a simulated case study for its practical use. It would be much better to have FedValue implemented with the standard of the industry and test it in various real-life VFL applications. The feedback from practices can further improve FedValue greatly.

[9]https://www.gartner.com/en/newsroom/press-releases/2021-09-30-gartner-says-digital-ethics-is-at-the-peak-of-inflate

# APPENDICES

## A DERIVATION OF $\varphi_d$ IN EXAMPLE 1.

Let's utilize the Shapley value defined in Eq. (1) to evaluate a data party $d$ in a cooperative game. We first enumerate the potential coalition $S$ formed before $d$'s join. In Example 1, there are only two parties (*i.e.*, the data party $d$ and the task party $t$); then, $d$ would be either the first party (*i.e.*, $S = \emptyset$) or the second party after $t$ (*i.e.*, $S = \{t\}$) joining the game. According to the definition of Shapley value, we compute the value of data party $d$ by its expected marginal contribution to the game as:

$$\varphi_d = \sum_{S \subseteq (\mathbb{P} \setminus \{d\})} \frac{|S|!(|\mathbb{P}| - |S| - 1)!}{|\mathbb{P}|!} (v(S \cup \{d\}) - v(S)) \tag{17}$$

$$= \frac{0!(2-0-1)!}{2!}(v(\{d\}) - v(\emptyset)) + \frac{1!(2-1-1)!}{2!}(v(\{d,t\}) - v(\{t\})) \tag{18}$$

$$= \frac{v(\{d\})}{2} + \frac{v(\{d,t\})}{2} - \frac{v(\{t\})}{2} \tag{19}$$

$$= \frac{v(\{d,t\})}{2} = \frac{v(\boldsymbol{X}_d)}{2} = \frac{v(\boldsymbol{X}_t)}{2}, \tag{20}$$

where $\frac{v(\{d\})}{2} = \frac{v(\{t\})}{2}$ as it is assumed the data party and the task party exhibit the same feature matrix (*i.e.*, $\boldsymbol{X}_d = \boldsymbol{X}_t$) in Example 1.

## B PROOF OF PROPOSITION 1

**Proof of 'Master-aware Efficiency', 'Symmetry', and 'Null Player'**

The first three properties can be naturally inferred from the original properties (efficiency, symmetry, and null player) of the Shapley value. Note that the task party's MShpaley-CMI value $\varphi_t (= I(\boldsymbol{X}_t; Y_t))$ is only dependent on the task party's own data set. If we only see data parties $\boldsymbol{\mathcal{D}}$ as players, then they form a traditional cooperative game with the total value as,

$$v(\boldsymbol{\mathcal{D}}) = I(\boldsymbol{X}_{\boldsymbol{\mathcal{D}}}; Y_t | \boldsymbol{X}_t) = I(\boldsymbol{X}_{\boldsymbol{\mathcal{D}}} \boldsymbol{X}_t; Y_t) - I(\boldsymbol{X}_t; Y_t) \tag{21}$$

Hence, the first three properties can be easily deduced from the Shapley value properties.

**Proof of 'Null Duplication'**

For proving the fourth property, consider the entropy computation $H(Y_t|\boldsymbol{X}_S\boldsymbol{X}_d)$ $(d \notin S)$:

$$H(Y_t|\boldsymbol{X}_S\boldsymbol{X}_d) = -\sum_{x_S \in \mathcal{X}_S} p(x_S x_d) \sum_{y_t \in \mathcal{Y}_t} p(y_t|x_S x_d) \log p(y_t|x_S x_d), \tag{22}$$

By duplicating certain features in $d$, $x_d = \langle x_d^{(1)}, x_d^{(2)} \rangle$ is changed to $\tilde{x}_d = \langle x_d^{(1)}, x_d^{(2)}, x_d^{(2)} \rangle$. As the duplicated feature is always same as $x_d^{(2)}$, we have:

$$p(x_S x_d) = p(x_S \tilde{x}_d), \quad p(y_t|x_S x_d) = p(y_t|x_S \tilde{x}_d) \tag{23}$$

Then,

$$H(Y_t|\boldsymbol{X}_S\boldsymbol{X}_d) = H(Y_t|\boldsymbol{X}_S\widetilde{\boldsymbol{X}}_d) \tag{24}$$

Hence, the characteristic function $v_{y_t}(S \cup \{d\})$ keeps unchanged after $d$ duplicates features, and then MShapley-CMI is also unchanged. □

## C PROOF OF PROPOSITION 2

According to Definition 5 and Eq. 13, to calculate an MShapley-CMI value, we need to compute cardinality $N(\cdot)$ for each possible combination of $\boldsymbol{x}_d$, $\boldsymbol{x}_D$, $\boldsymbol{x}_t$, and $y_t$. Let $\mathcal{Y}_t$, $\mathcal{X}_t$, and $\mathcal{X}_d$ $(\forall d \in \boldsymbol{\mathcal{D}})$ denote the sets of all possible values of $Y_t$, $\boldsymbol{X}_t$, and $\boldsymbol{X}_d$, respectively. Therefore, We need to compute cardinality $N(\cdot)$ for $|\mathcal{Y}_t||\mathcal{X}_t| \cdot \Pi_{d \in \boldsymbol{\mathcal{D}}}|\mathcal{X}_d|$ times.

Let us consider the computation of one $N(\cdot)$ by invoking Algorithm 1. The probability that a falsified $N(\cdot)$ by a malicious server passes the divisibility and non-negativity verifications in one iteration is no larger than $\min(q_{\text{ID}})^{-1}$, where $\min(q_{\text{ID}})$ represents the minimum number of ID duplications across iterations. Considering that there are $\tau$ iterations, the probability of a falsified $N(\cdot)$ passing all the verifications in Algorithm 1 is no larger than $\min(q_{\text{ID}})^{-\tau}$. Given that there are $|\mathcal{Y}_t||\mathcal{X}_t| \cdot \Pi_{d \in \boldsymbol{\mathcal{D}}}|\mathcal{X}_d|$ cardinality computations, the probability that a malicious server can successfully deceive the task party during the process of computing an MShapley-CMI value is no larger than $\min(q_{\text{ID}})^{-\tau|\mathcal{Y}_t||\mathcal{X}_t| \cdot \Pi_{d \in \boldsymbol{\mathcal{D}}}|\mathcal{X}_d|}$.

□

**Table 8. Well-performing model selection by prediction accuracy. For each dataset, the accuracy of the best model is in bold; the models with a double-underlined accuracy would be selected when threshold $\epsilon = 0.02$; all the models with an (both single- and double-) underlined accuracy would be selected when threshold $\epsilon = 0.05$. For example, in the *Parkinsons* dataset, the SVM, GBT, and RF models are selected into $M_{well}$ when $\epsilon$ is set to 0.05.**

| | | *SVM* | *GBT* | *LR* | *RF* | *NN* |
|---|---|---|---|---|---|---|
| *Wine* | accuracy | **0.967** | 0.944 | **0.967** | 0.944 | **0.967** |
| | Std. | ±0.027 | ±0.030 | ±0.011 | ±0.025 | ±0.041 |
| *Parkinsons* | accuracy | 0.897 | 0.908 | 0.821 | **0.928** | 0.856 |
| | Std. | ±0.043 | ±0.053 | ±0.036 | ±0.030 | ±0.058 |
| *Spect* | accuracy | **0.833** | 0.819 | 0.819 | 0.822 | 0.826 |
| | Std. | ±0.012 | ±0.032 | ±0.041 | ±0.022 | ±0.045 |
| *Breast* | accuracy | 0.951 | 0.937 | **0.961** | 0.937 | 0.949 |
| | Std. | ±0.012 | ±0.020 | ±0.007 | ±0.013 | ±0.010 |
| *Music* | accuracy | 0.605 | 0.643 | 0.596 | **0.653** | 0.613 |
| | Std. | ±0.021 | ±0.016 | ±0.034 | ±0.024 | ±0.030 |
| *Credit* | accuracy | 0.809 | **0.812** | 0.808 | 0.795 | 0.805 |
| | Std. | ±0.004 | ±0.005 | ±0.006 | ±0.006 | ±0.005 |

## D EXPERIMENTAL SETUPS AND ADDITIONAL RESULTS

To compare with model-dependent data valuation methods, we first train a set of prediction models with various machine learning methods, including support vector machine (SVM), gradient boosting decision tree (GBT), logistic regression (LR), random forest (RF), and neural networks (NN). Table 8 shows the prediction accuracy of different models on various datasets. Let $acc^*$ denote the best model's accuracy and $\epsilon$ denote a threshold for model selection. For each dataset, we select the models with a prediction accuracy larger than $(acc^* - \epsilon)$ as the well-performing ones (denoted as $M_{well}$). $\epsilon$ is set to $0.02$ and $0.05$ respectively in the experiments. Then, given all well-performing models in $M_{well}$, we use a model-dependent feature importance metric to valuate each party's aggregated feature and average the valuation results by different models as reference.

SHAP (Lundberg and Lee, 2017), Permutation Importance (PI) (Altmann et al., 2010) and LIME (Ribeiro et al., 2016) are the widely used model-dependent feature importance metrics. In general, PI cannot appropriately valuate the data parties that share some features[10]. Compartively, LIME is mostly used at an individual sample level (i.e., explain why a sample is predicted as a certain label) instead of a global model level, and SHAP can be seen as an improved version of LIME (especially the Kernel SHAP algorithm) to provide the better global interpretability (Lundberg and Lee, 2017). As a result, we ensemble SHAP values of a party's aggregated feature given the well-performing models as a reference of the party's value, denoted as *SHAP-ensemble*.

[10] A detailed discussion can be found in Sec. 5.6 of *Interpretable Machine Learning* (`https://christophm.github.io/interpretable-ml-book/feature-importance.html`).

## E ROBUSTNESS CHECK ON THE APPLICATION OF FEDVALUE FOR FEDERATED RECOMMENDATION

Given a default budget of 2 and the FIX incentive scheme, we conduct the robustness check by varying numerous parameters used in federated recommendation, including the number of movies for recommendation, the number of data party candidates, the feature vector width, the data ratios of high-volume data parties, the redundant candidate ratio, as well as the free-rider candidate ratio. In particular, we use the varying values of one particular parameter and the default values for the other parameters at a time to obtain the results. The experimental results reveal that FedValue outperforms VOLUME, RANDOM, and FL-PS consistently cross all the robustness check parameter settings.

**Table 9. Robustness check.**

| **Parameter** | | **accuracy** | | | | **AUC** | | | |
|---|---|---|---|---|---|---|---|---|---|
| name | value | RANDOM | VOLUME | VF-PS | FedValue | RANDOM | VOLUME | VF-PS | FedValue |
| recommended movies | 30 | 0.724 | 0.722 | 0.727 | **0.729** | 0.758 | 0.742 | 0.759 | **0.760** |
| | 40 | 0.707 | 0.699 | 0.711 | **0.714**** | 0.760 | 0.740 | 0.759 | **0.763*** |
| data party candidates | 4 | 0.744 | 0.746 | 0.747 | **0.748** | 0.739 | 0.753 | 0.753 | **0.755** |
| | 6 | 0.743 | 0.743 | 0.745 | **0.747*** | 0.751 | 0.741 | 0.757 | **0.761*** |
| feature vector width | 1 | 0.736 | 0.735 | 0.739 | **0.743**** | 0.716 | 0.702 | 0.718 | **0.721*** |
| | 3 | 0.751 | 0.756 | 0.752 | **0.759**** | 0.776 | 0.778 | 0.777 | **0.784**** |
| high-volume data ratio | 5% | 0.741 | 0.742 | 0.744 | **0.746** | 0.754 | 0.747 | 0.758 | **0.761** |
| | 20% | 0.744 | 0.743 | 0.746 | **0.749**** | 0.751 | 0.731 | 0.753 | **0.755** |
| redundant candidate ratio | 40% | 0.717 | 0.718 | 0.720 | **0.737***** | 0.736 | 0.729 | 0.744 | **0.767***** |
| | 60% | 0.715 | 0.715 | 0.721 | **0.741***** | 0.711 | 0.706 | 0.711 | **0.761***** |
| free-rider candidate ratio | 40% | 0.720 | 0.719 | 0.743 | **0.745*** | 0.711 | 0.702 | 0.758 | **0.764***** |
| | 60% | 0.693 | 0.685 | 0.740 | **0.743** | 0.659 | 0.645 | 0.746 | **0.753**** |